%% file: main_arxiv.tex
\documentclass[runningheads]{llncs}
\usepackage{graphicx}

\usepackage{tikz}
\usepackage{comment}
\usepackage{amsmath,amssymb} %
\usepackage{color}
\usepackage{algorithm}
\usepackage{algpseudocode}

\usepackage{subcaption} %
\usepackage{pgfplots}  %
\usepackage{adjustbox} %

\usepackage[accsupp]{axessibility}  %

\DeclareMathOperator*{\argmin}{argmin}

\usepackage{overpic}
\usepackage{enumitem} %
\usepackage{overpic} %
\usepackage{xcolor}
\usepackage{booktabs}
\usepackage{flushend}
\usepackage{soul}
\setstcolor{red}

\usepackage[pagebackref,breaklinks,colorlinks,citecolor=blue]{hyperref}

\newcommand{\parag}[1]{\smallskip\noindent\textbf{#1}}

\begin{document}
\pagestyle{headings}
\mainmatter
\def\ECCVSubNumber{4953}  %

\newcommand{\todo}[1]{\textcolor{purple}{TODO: #1}}
\newcommand{\cccv}[1]{\textcolor{brown}{#1}}
\newcommand{\ie}{i.e.,\ }
\newcommand{\eg}{e.g.,\ }

\newcommand{\acro}{STEEX}
\newcommand{\targetedce}{region-targeted counterfactual explanations}
\newcommand{\Targetedce}{region-targeted counterfactual explanations}
\newcommand{\dset}[1]{\textsf{#1}}

\title{STEEX: Steering Counterfactual Explanations with Semantics}

\institute{Paper ID \ECCVSubNumber}

\titlerunning{STEEX: Steering Counterfactual Explanations with Semantics}
\author{Paul Jacob \inst{1} \index{Jacob, Paul} \and
\'Eloi Zablocki \inst{1} \index{Zablocki, \'Eloi} \and
Hédi Ben-Younes \inst{1} \index{Ben-Younes, Hédi} \and
Mickaël Chen \inst{1} \index{Chen, Mickaël} \and
Patrick Pérez \inst{1} \index{Pérez, Patrick} \and
Matthieu Cord \index{Cord, Matthieu} \inst{1,2}
}
\authorrunning{P. Jacob et al.}
\institute{Valeo.ai \and Sorbonne University
}

\maketitle

\begin{abstract}
As deep learning models are increasingly used in safety-critical applications, explainability and trustworthiness become major concerns. For simple images, such as low-resolution face portraits, synthesizing visual counterfactual explanations has recently been proposed as a way to uncover the decision mechanisms of a trained classification model. 
In this work, we address the problem of producing counterfactual explanations for high-quality images and complex scenes. Leveraging recent semantic-to-image models, we propose a new generative counterfactual explanation framework that produces plausible and sparse modifications which preserve the overall scene structure.
Furthermore, we introduce the concept of ``region-targeted counterfactual explanations'', and a corresponding framework, where users can guide the generation of counterfactuals by specifying a set of semantic regions of the query image the explanation must be about.
Extensive experiments are conducted on challenging datasets including high-quality portraits (\dset{CelebAMask-HQ}) and driving scenes (\dset{BDD100k}). Code is available at: \url{https://github.com/valeoai/STEEX}

\end{abstract}

\section{Introduction}

\input{bigpicture.tex}

Deep learning models are now used in a wide variety of application domains, including safety-critical ones. 
As the underlying mechanisms of these models remain opaque, explainability and trustworthiness have become major concerns. 
In computer vision, \emph{post-hoc} explainability often amounts to producing saliency maps, which highlight regions on which the model grounded the most its decision \cite{cam,gradcam,lrp,deeplift,integrated_gradients,deconvnet,meaningful_perturbation}.
While these explanations show \emph{where} the regions of interest for the model are, they fail to indicate \emph{what} specifically in these regions leads to the obtained output.
A desirable explanation should not only be \emph{region}-based but also \emph{content}-based by expressing in some way how the content of a region influences the outcome of the model. 
For example, in autonomous driving, while it is useful to know that a stopped self-driving car attended the traffic light, it is paramount to know that the red color of the light was decisive in the process.

In the context of simple tabular data, \emph{counterfactual explanations} have recently been introduced to provide fine content-based insights on a model's decision \cite{wachter2017counterfactual,counterfactual_explanations_review,explaining_classifiers_counterfactual}.
Given an input query, a counterfactual explanation is a version of the input with \emph{minimal} but \emph{meaningful} modifications that change the output decision of the model.
\emph{Minimal} means that the new input must be as similar as possible to the query input, with only sparse changes or in the sense of some distance to be defined. 
\emph{Meaningful} implies that changes must be semantic, \ie human-interpretable.
This way, a counterfactual explanation points out in an understandable way \emph{what} is important for the decision of the model by presenting a close hypothetical reality that contradicts the observed decision. 
As they are contrastive and as they usually focus on a small number of feature changes, counterfactuals can increase user's trust in the model \cite{shen2020explain,trusthci20,xai_driving_survey}.
Moreover, these explanations can also be leveraged by machine learning engineers, as they can help to identify spurious correlations captured by a model \cite{deeptest,deeproad,dive}.
Despite growing interest, producing visual counterfactual explanations for an image classification model is especially challenging as naively searching for small input changes results in adversarial perturbations \cite{adversarial2014,adversarial2015,ce_ae_common_grounds,connections_ce_ae,why_ce_produce_ae}.
To this date, there only exists a very limited number of counterfactual explanation methods able to deal with image classifiers \cite{counterfactual_visual_explanations,scout,progressive_exaggeration,dive}.
Yet, these models present significant limitations, as they either require a target image of the counterfactual class \cite{counterfactual_visual_explanations,scout} or can only deal with classification settings manipulating simple images such as low-resolution face portraits \cite{progressive_exaggeration,dive}.

In this work, we tackle the generation of counterfactual explanations for deep classifiers operating on large images and/or visual scenes with complex structures. 
Dealing with such images comes with unique challenges, beyond technical issues.
Indeed, because of scene complexity, it is likely that the model's decision can be changed by many admissible modifications in the input.
For a driving action classifier, it could be for instance modifying the color of traffic lights, the road markings or the visibility conditions, but also adding new elements to the scene such as pedestrians and traffic lights, or even replacing a car on the road with an obstacle.
Even if it was feasible to provide an exhaustive list of counterfactual explanations, the task of selecting which ones in this large collection are relevant would fall on the end-user, hindering the usability of the method.
To limit the space of possible explanations while preserving sufficient expressivity, we propose that the overall structure of the query image remains untouched when creating the counterfactual example.
Accordingly, through semantic guidance, we impose that a generated counterfactual explanation respects the original layout of the query image.

Our model, called \acro{} for STEering counterfactual EXplanations with semantics, leverages recent breakthroughs in semantic-to-real image synthesis \cite{oasis,spade,sean}. 
A pre-trained encoder network decomposes the query image into a spatial layout structure and latent representations encoding the content of each semantic region.
By carefully modifying the latent codes towards a different decision, \acro{} is able to generate meaningful counterfactuals with relevant semantic changes and a preserved scene layout, as illustrated in \autoref{fig:bigpicture:b}.
Additionally, we introduce a new setting where users can guide the generation of counterfactuals by specifying which semantic region of the query image the explanation must be about.
We coin ``\targetedce'' such generated explanations where only a subset of latent codes is allowed to be modified.
In other words, such explanations are answers to questions such as \emph{``How should the traffic lights change to switch the model's decision?''}, as illustrated in \autoref{fig:bigpicture:d}.
To validate our claims, extensive experiments of \acro{} are conducted on a variety of image classification models trained for different tasks, including self-driving action decision on the \dset{BDD100k} dataset, and high-quality face recognition networks trained on \dset{CelebAMask-HQ}.
Besides, we investigate how explanations for different decision models can hint at their distinct and specific behaviors.

To sum up, our contributions are as follows:
\begin{itemize}[leftmargin=6pt,topsep=2pt,itemsep=0pt,parsep=0pt,partopsep=0pt]
    \item We tackle the generation of visual counterfactual explanations for classifiers dealing with large and/or complex images.
    \item By leveraging recent semantic-to-image generative models, we propose a new framework capable of generating counterfactual explanations that preserve the semantic layout of the image.
    \item We introduce the concept of ``region-targeted counterfactual explanations'' %
    to target specified semantic regions in the counterfactual generation process.
    \item We validate the quality, plausibility and proximity to their query, of obtained explanations with extensive experiments, including classification models for high-quality face portraits and complex urban scenes.
\end{itemize}

\section{Related Work}
\label{sec:rw}

The black-box nature of deep neural networks has led to the recent development of many explanation methods  \cite{beaudouin2020identifying,GilpinBYBSK18,xai_peeking_inside_black_box,opportunities_and_challenges_in_xai_survey}.
In particular, our work is grounded within the \emph{post-hoc} explainability literature aiming at explaining a trained model, which contrasts with approaches building interpretable models \emph{by design} \cite{interpretable_cnn,this_looks_like_that}.
Post-hoc methods can be either be \emph{global} if they seek to explain the model in its entirety, or \emph{local} when they explain the prediction of the model for a specific instance.
Global approaches include model translation techniques, that distill the black-box model into a more interpretable one \cite{soft_decision_tree,causal_learning_autoencoded_activations}, or the more recent disentanglement methods that search for latent dimensions of the input space that are, at the level of the dataset, correlated with the output variations of the target classifier \cite{stylex,discoverbiasedattr2021}.
Instead, in this paper, we focus on \emph{local} methods that provide explanations, tailored to a given image.

Usually, \emph{post-hoc} \emph{local} explanations of vision models are given in the form of saliency maps, which attribute the output decision to image regions.
Gradient-based approaches compute this attribution using the gradient of the output with respect to input pixels or intermediate layers \cite{gradcam,integrated_gradients,there_and_back_again,visualbackprop}.
Differently, perturbation-based approaches  \cite{deconvnet,object_detectors_emerge_in_deep_scene,meaningful_perturbation,interpretable_fine_grained_expl} evaluate how sensitive to input variations is the prediction.
Other explainability methods include locally fitting a more interpretable model such as a linear function \cite{lime} or measuring the effect of including a feature with game theory tools \cite{shap}.
However, these methods only provide information on \emph{where} are the regions of interest for the model but do not tell \emph{what} in these regions is responsible for the decision.

Counterfactual explanations \cite{wachter2017counterfactual}, on the other hand, aim to inform a user on why a model $M$ classifies a specific input $x$ into class $y$ instead of a \emph{counter class} $y' \neq y$. 
To do so, a \emph{counterfactual example} $x'$ is constructed to be similar to $x$ but classified as $y'$ by $M$. 
Seminal methods have been developed in the context of low-dimensional input spaces, like the ones involved in credit scoring tasks \cite{wachter2017counterfactual}.
Naive attempts to scale the concept to higher-dimensional input spaces, such as natural images, face the problem of producing adversarial examples \cite{adversarial2014,adversarial2015,deepfool,why_ce_produce_ae}, that is, \textit{imperceptible} changes to the query image that switch the decision.
While the two problems have similar formulations, their goals are in opposition \cite{ce_ae_common_grounds,connections_ce_ae} since counterfactual explanations must be understandable, achievable, and informative for a human.
Initial attempts to counterfactual explanations of vision models would explain a decision by comparing the image $x$ to one or several real instances classified as $y'$ \cite{grounding_visual_explanations,counterfactual_visual_explanations,scout}. 
However, these discriminative counterfactuals do not produce natural images as explanations, and their interpretability is limited when many elements vary from one image to another.

To tackle these issues, generative methods leverage deep generative models to produce counterfactual explanations.
For instance, DiVE \cite{dive} is built on $\beta$-TCVAE \cite{beta_tcvae} and takes advantage of its disentangled latent space to discover such meaningful sparse modifications.
With this method, it is also possible to generate multiple orthogonal changes that correspond to different valid counterfactual examples.
Progressive Exaggeration (PE) \cite{progressive_exaggeration}, instead, relies on a Generative Adversarial Network (GAN) \cite{gan2016} conditioned on a perturbation value that is introduced as input in the generator via conditional batch normalization.
PE modifies the query image so that the prediction of the decision model is shifted by this perturbation value towards the counter class.
By applying this modification multiple times, and by showing the progression, PE highlights adjustments that would change the decision model's output.
Unfortunately, none of these previous works is designed to handle complex scenes. 
The $\beta$-TCVAE used in DiVE hardly scales beyond small centered images, requiring specifically-designed enhancement methods \cite{high_fidelity_disentangled,improving_reconstruction_disentangled}, and PE performs style-based manipulations that are unsuited images with multiple small independent objects of interest.
Instead, our method relies on segmentation-to-image GANs \cite{spade,oasis,sean}, that have demonstrated good generative capabilities on high-quality images containing multiple objects.

\section{Model \acro{}}
\label{sec:model}

We now describe our method to obtain counterfactual explanations with semantic guidance.
First, we formalize the generative approach for visual counterfactual explanations in \autoref{sec:model:generative_high_level}.
Within this framework, we then incorporate a semantic guidance constraint in \autoref{sec:model:semantic_guided}.
Next we propose in \autoref{sec:model:region_targeted} a new setting where the generation targets specified semantic regions.
Finally, \autoref{sec:model:model_instantiation} details the instantiation of each component.
An overview of \acro{} is presented in \autoref{fig:architecture}.

\subsection{Visual Counterfactual Explanations}
\label{sec:model:generative_high_level}

Consider a trained differentiable machine learning model $M$, which takes an image $x^I \in \mathcal{X}$ from an input space $\mathcal{X}$ and outputs a prediction $y^I = M(x^I) \in \mathcal{Y}$.
A counterfactual explanation for the obtained decision $y^I$ is an image $x$ which is as close to the image $x^I$ as possible, but such that $M(x) = y$ where $y \neq y^I$ is another class.
This problem can be formalized and relaxed as follows:
\begin{equation}
    \argmin\nolimits_{x \in \mathcal{X}} L_{\text{decision}}(M(x), y) + \lambda L_{\text{dist}}(x^I,x) ,
\end{equation}
where $L_{\text{decision}}$ is a classification loss,
$L_{\text{dist}}$ measures the distance between images, and the hyperparameter $\lambda$ balances the contribution of the two terms. 

In computer vision applications where input spaces are high-dimensional, additional precautions need to be taken to avoid ending up with adversarial examples \cite{adversarial2014,ce_ae_common_grounds,connections_ce_ae,why_ce_produce_ae}.
To prevent those uninterpretable perturbations, which leave the data manifold by adding imperceptible high-frequency patterns, counterfactual methods impose that visual explanations lie in the original input domain $\mathcal{X}$. 
Incorporating this in-domain constraint can be achieved by using a deep generator network as an implicit prior \cite{deep_image_prior,csgan}.
Consider a generator $G:z \mapsto x$ that maps vectors $z$ in latent space $\mathcal{Z}$ to in-distribution images $x$.
Searching images only in the output space of such a generator would be sufficient to satisfy the in-domain constraint, and the problem now reads:
\begin{equation}
    \label{eq:counterfactual_generative}
    \argmin\nolimits_{z \in \mathcal{Z}} L_{\text{decision}}(M(G(z)), y) + \lambda L_{\text{dist}}(x^I,G(z)) .
\end{equation}
\autoref{eq:counterfactual_generative} formalizes practices introduced in prior works \cite{dive,progressive_exaggeration} that also aim to synthesize counterfactual explanations for images.

Furthermore, assuming that a latent code $z^I$ exists and can be recovered for the image $x^I$, we can express the distance loss directly in the latent space $\mathcal{Z}$:
\begin{equation}
    \label{eq:counterfactual_latent}
    \argmin\nolimits_{z \in \mathcal{Z}}  L_{\text{decision}}(M(G(z)), y) + \lambda L_{\text{dist}}(z^I,z) .
\end{equation}
By searching for an optimum in a low-dimensional latent space rather than in the raw pixel space, we operate over inputs that have a higher-level meaning, which is reflected in the resulting counterfactual examples.

\input{overview.tex}

\subsection{Semantic-Guided Counterfactual Generation}
\label{sec:model:semantic_guided}

The main objective of our model is to scale counterfactual image synthesis to large and complex scenes involving multiple objects within varied layouts. 
In such a setting, identifying and interpreting the modifications made to the query image is a hurdle to the usability of counterfactual methods.
Therefore we propose to generate counterfactual examples that preserve the overall structure of the query and, accordingly, design a framework that optimizes under a fixed semantic layout.
Introducing semantic masks for counterfactual explanations comes with additional advantages.
First, we can leverage semantic-synthesis GANs that are particularly well-suited to generate diverse complex scenes \cite{spade,sean,oasis}.
Second, it provides more control over the counterfactual explanation we wish to synthesize, allowing us to target the changes to a specific set of semantic regions, as we detail in \autoref{sec:model:region_targeted}.
To do so, we adapt the generator $G$ and condition it on a semantic mask $S$ that associates each pixel to a label indicating its semantic category (for instance, in the case of a driving scene, such labels can be cars, road, traffic signs, etc.). The output of the generator ${G:(S,z) \mapsto x}$ is now restricted to follow the layout indicated by $S$.
We can then find a counterfactual example for image $x^I$ that has an associated semantic mask $S^I$ by optimizing the following objective:
\begin{equation}
    \label{eq:semantic-conterfactual}
    \argmin\nolimits_{z\in\mathcal{Z}}  L_{\text{decision}}(M(G(S^I, z)), y) + \lambda L_{\text{dist}}(z^I,z) . 
\end{equation}
This formulation guarantees that the semantic mask $S^I$ of the original scene is kept as is in the counterfactuals.

\subsection{Region-Targeted Counterfactual Explanations}
\label{sec:model:region_targeted}

We introduce a new setting enabling finer control in the generation of counterfactuals.
In this setup, a user specifies a set of semantic regions that the explanation must be about.
For example, in \autoref{fig:architecture}, the user selects `car' and `traffic light', and the resulting counterfactual is only allowed to alter these regions. Such a selection allows studying the influence of different semantic concepts in the image for the target model's behavior.
In practice, given a semantic mask $S$ with $N$ classes, we propose to decompose $z$ into $N$ vectors, $z = (z_c)_{c=1}^{N}$, where each $z_c$ is a latent vector associated with one class in $S$.
With such a formulation, it becomes possible to target a subset $C\subset\{1,\dots,N\}$  for the counterfactual explanation.
Region-targeted counterfactuals only optimize on the specified components 
$(z_c)_{c\in C}$, %
and all other latent codes remain unmodified.

\subsection{Instantiation of \acro{}}
\label{sec:model:model_instantiation}

We now present the modeling choices we make for each part of our framework. 

\parag{Generator $G$.}
The generator $G$ can be  %
any of the recent segmentation-to-image GANs \cite{spade,sean,oasis} that transform a latent code $z$ and a segmentation layout $S$ into an image $x$.
As such generators typically allow for a different vector $z_c$ to be used for each class in the semantic mask \cite{sean,oasis}, the different semantic regions can be modified independently in the output image.
This property enables \acro{} to perform region-targeted counterfactual explanations as detailed in \autoref{sec:model:region_targeted}.

\parag{Obtaining the code $z^I$.}
To recover the latent code $z^I$ from the image $x^I$, we exploit the fact that in aforementioned frameworks \cite{sean,oasis}, the generator $G$ can be trained jointly, in an auto-encoding pipeline, with an encoder $E_{z}$ that maps an image $x^I$ and its associated segmentation layout $S^I$  into a latent code $z^I$. Such a property ensures that we can efficiently compute this image-to-latent mapping and that there is indeed a semantic code that corresponds to each image, leading to an accurate reconstruction in the first place.

\parag{Obtaining the mask $S^I$.}
As query images generally have no associated annotated segmentation masks $S^I$, these need to be inferred.
To do so, we add a segmentation network $E_\text{seg}$ in the pipeline: we first obtain the map ${S^I\,{=}\, E_\text{seg}(x^I)}$ and then use the encoder: ${z^I\,{=}\, E_{z}(x^I, S^I)}$, so \acro{} is applicable to any image.

\parag{Loss functions.}
The decision loss $L_\text{dist}$ ensures that the output image $x$ is classified as $y$ by the decision model $M$. It is thus set as the negative log-likelihood of the targeted counter class $y$ for $M(G(z))$:
\begin{equation}
L_{\text{decision}}(M(G(z)), y) = -\mathcal{L}(M(G(z))|y) .
\end{equation}
The distance loss $L_\text{dist}$ is the sum of squared L2 distance between each semantic component of $z^I$ and $z$:
\begin{equation}
    L_{\text{dist}}(z^I,z) = \sum\limits_{c=1}^{N} \| z_c^I - z_c \|^2_2 .
\end{equation}
We stress that \autoref{eq:semantic-conterfactual} is optimized on the code $z$ only. All of the network parameters ($G$, $E_z$ and $E_\text{seg}$) remain frozen.

\section{Experiments}

We detail in \autoref{sec:expe:protocol} our experimental protocol to evaluate different aspects of generated counterfactuals: the plausibility and perceptual quality (\autoref{sec:expe:perceptual_quality}) as well as the proximity to query images (\autoref{sec:expe:expe_sparsity}).
We then present in \autoref{sec:expe:region_targeted} region-targeted counterfactual explanations.
In \autoref{sec:expe:analyzing_decision_models}, we use STEEX to explain different decision models for the same task, and show that produced explanations hint at the specificities of each model. 
Finally, we present an ablation study in \autoref{sec:expe:ablation}.
Our code and pretrained models will be made available.

\subsection{Experimental Protocol}
\label{sec:expe:protocol}

We evaluate our method on five decision models across three different datasets. We compare against two recently proposed visual counterfactual generation frameworks, Progressive Exaggeration (PE) \cite{progressive_exaggeration} and DiVE \cite{dive}, previously introduced in \autoref{sec:rw}.
We report scores directly from their paper when available (\dset{CelebA}) and used the public and official implementation to evaluate them otherwise (\dset{CelebAMask-HQ} and \dset{BDD100k}). 
We now present each dataset and the associated experimental setup.

\parag{\dset{BDD100k} \cite{bdd}.}
The ability of \acro{} to explain models handling complex visual scenes is evaluated on the driving scenes of \dset{BDD100k}. Most images of this dataset contain diversely-positioned objects that can have fine relationships with each other, and small details in size can be crucial for the global understanding of the scene (\eg traffic light colors).
The decision model to be explained is a \emph{Move Forward} vs.\ \emph{Stop/Slow down} action classifier trained on \dset{BDD-OIA} \cite{bdd_oia}, a 20,000-scene extension of \dset{BDD100k} annotated with binary attributes representing the high-level actions that are allowed in a given situation. The image resolution is $512\times256$.
The segmentation model $E_\text{seg}$ is a DeepLabV3 \cite{deeplabv3} trained on a subset of 10,000 images annotated with semantic masks that cover 20 classes (\eg road, truck, car, tree, etc.).
On the same set, the semantic encoder $E_z$ and the generator $G$ are jointly trained within a SEAN framework \cite{sean}.
Counterfactual scores are computed on the validation set of \dset{BDD100k}.

\parag{\dset{CelebAMask-HQ} \cite{CelebAMask-HQ}.}
\dset{CelebAMask-HQ} contains 30,000 high-quality face portraits with semantic segmentation annotation maps including 19 semantic classes (\eg skin, mouth nose, etc.).
The portraits are also annotated with identity and 40 binary attributes, allowing us to perform a quantitative evaluation for high-quality images. 
Decision models to be explained are two DenseNet121 \cite{densenet} binary classifiers trained to respectively recognize \emph{Smile} and \emph{Young} attributes.
To obtain semantic segmentation masks for the query images, we instantiate $E_\text{seg}$ with a DeepLabV3 \cite{deeplabv3} pre-trained on the 28,000-image training split.
On the same split, the semantic encoder $E_z$ and generator $G$ are jointly learned within a SEAN framework \cite{sean}.
Counterfactual explanations are computed on the 2000-image validation set, with images rescaled to the resolution $256\times256$.

\parag{\dset{CelebA} \cite{celeba}.}
\dset{CelebA} contains 200,000 face portraits, annotated with identity and 40 binary attributes, but of smaller resolution ($128 \times 128$ after processing) and of lower quality compared to \dset{CelebAMask-HQ}.
\acro{} is designed to handle more complex and larger images, but we include this dataset for the sake of completeness as previous works \cite{progressive_exaggeration,dive} use it as their main benchmark. 
We report their score directly from their respective papers and align our experiment protocol with the one described in \cite{dive}.
As in previous works, we explain two decision models: a \emph{Smile} classifier and a \emph{Young} classifier, both with DenseNet121 architecture \cite{densenet}.
We obtain $E_\text{seg}$ with a DeepLabV3 \cite{deeplabv3} trained on \dset{CelebAMask-HQ} images.
Then, we jointly train the semantic encoder $E_z$ and generator $G$ with a SEAN architecture \cite{sean} on the training set of \dset{CelebA}.
Explanations are computed on the 19,868-image validation split of \dset{CelebA}.

\parag{Optimization scheme.}
As $M$ and $G$ are differentiable, we optimize $z$ using ADAM \cite{adam} with a learning rate $1\cdot 10^{-2}$ for 100 steps with $\lambda=0.3$. Hyperparameters have been found on the training splits of the datasets.

\subsection{Quality of the Counterfactual Explanations}
\label{sec:expe:perceptual_quality}
We first ensure that the success rate of \acro{}, \ie the fraction of explanations that are well classified into the counter class, is higher than 99.5\% for all of the five tested classifiers. 
Then, as \acro{}'s counterfactuals must be realistic and informative, we evaluate their perceptual quality.

\input{tables/fid_celebahq_bdd}

\begin{figure*}[t]
    \centering
    \includegraphics[width=\linewidth]{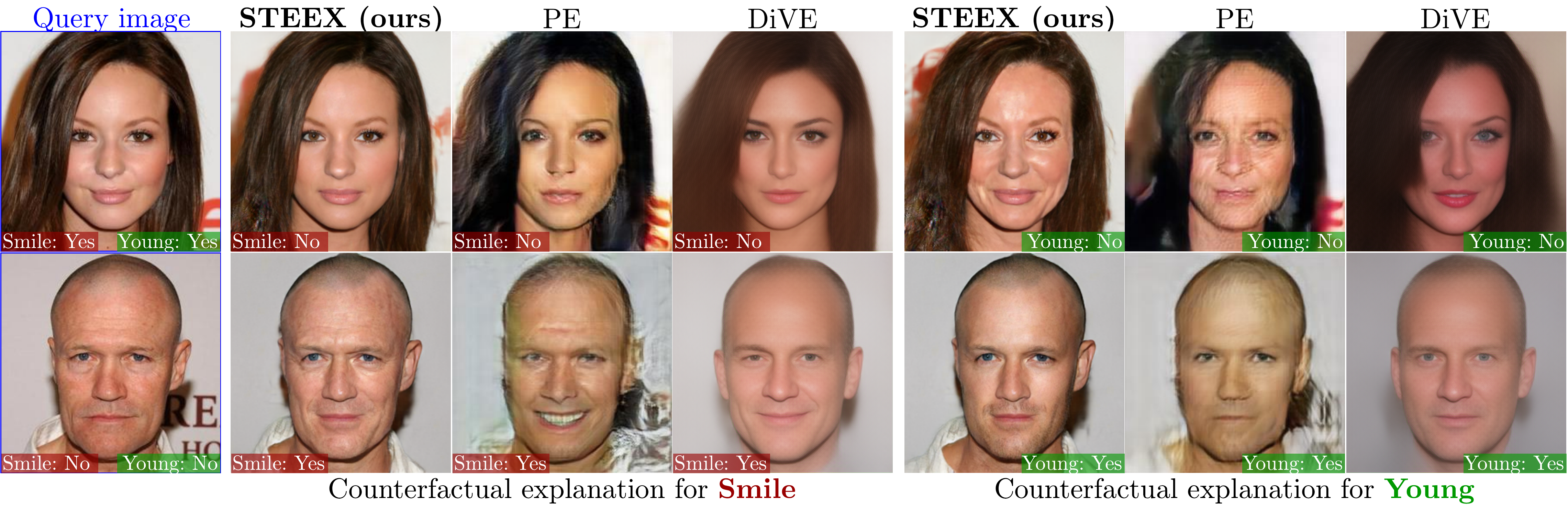}
    \caption{\textbf{Counterfactual explanations on \dset{CelebAMask-HQ}}, generated by \acro{} (ours), PE, and DiVE. Explanations are generated for two binary classifiers, on \emph{Smile} and \emph{Young} attributes, at resolution $256 \times 256$.
    Other examples in the Supplementary
    }
    \label{fig:celebamaskhq}
\end{figure*}

Similarly with previous works \cite{progressive_exaggeration,dive}, we use the Fréchet Inception Distance (FID) \cite{fid} between all explanations and the set of query images, and report this metric in \autoref{tab:FID_2}.
For each classifier, \acro{} outperforms the baselines by a large margin, meaning that our explanations are more realistic-looking, 
which verifies that they belong to the input domain of the decision model.

\begin{figure*}[b]
    \centering
    \includegraphics[width=\linewidth]{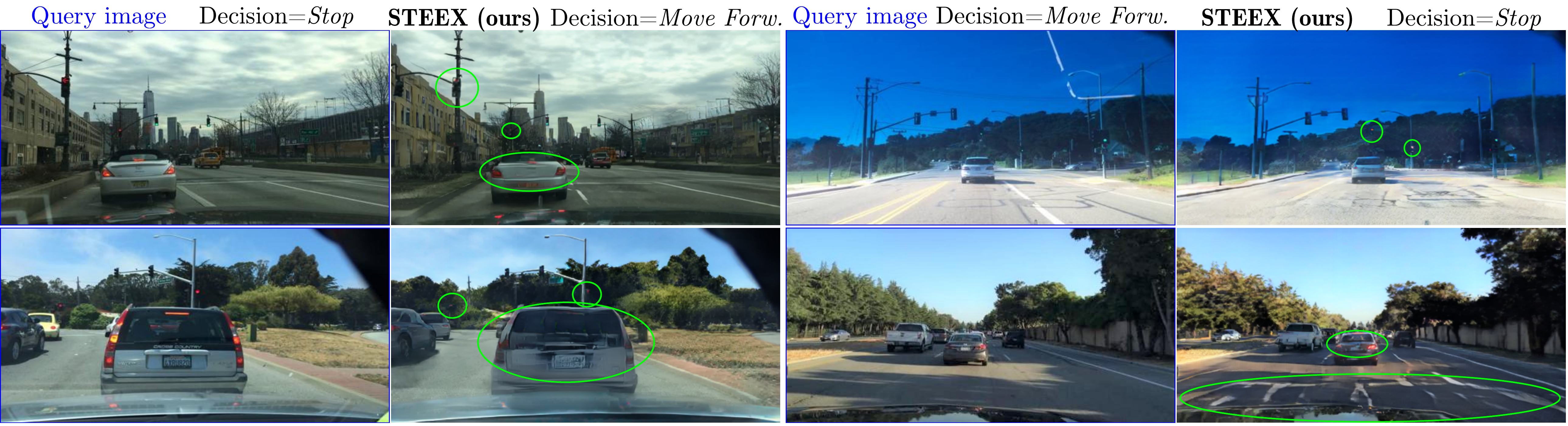}
    \caption{
    \textbf{Counterfactual explanations on \dset{BDD100k}.} Explanations are generated for a binary classifier for the action \emph{Move Forward}, with images at resolution $512 \times 256$. Our method finds interpretable, sparse and meaningful semantic modifications to the query image. 
    Other examples are available in the Supplementary
    }
    \label{fig:bdd_global}
\end{figure*}

Generating realistic counterfactuals for classifiers that deal with large and complex images is difficult, as reflected by large FID discrepancies between \dset{CelebA}, \dset{CelebAMask-HQ} and \dset{BDD100k}.
Scaling the generation of counterfactual explanations from $128 \times 128$ (\dset{CelebA}) to $256 \times 256$ (\dset{CelebAMask-HQ}) face portraits is not trivial as a significant drop in performance can be observed for all models, especially for DiVE.
Despite our best efforts to train DiVE on \dset{BDD100k}, we were unable to obtain satisfying $512 \times 256$ explanations, as all reconstructions were nearly uniformly gray. As detailed in \autoref{sec:rw}, VAE-based models are indeed usually limited to images with a fairly regular structure, and they struggle to deal with the diversity of driving scenes.

We display examples of \acro{}'s counterfactual explanations on \dset{CelebAMask-HQ} in \autoref{fig:celebamaskhq}, compared with PE \cite{progressive_exaggeration} and DiVE \cite{dive}. 
For the \emph{Smile} classifier, \acro{} explains positive (top-row) and negative (bottom-row) smile predictions through sparse and photo-realistic modifications of the lips and the skin around the mouth and the eyes. 
Similarly, for the \emph{Young} classifier, \acro{} explain decisions by adding or removing facial wrinkles. 
In comparison, PE introduces high-frequency artifacts that harm the realism of generated examples. 
DiVE generates blurred images and applies large modifications so that it becomes difficult to identify the most crucial changes for the target model.
\autoref{fig:bdd_global} shows other samples for the action classifier on the \dset{BDD100k} dataset, where we overlay green ellipses to point the reader's attention to significant region changes. 
\acro{} finds sparse but highly semantic modifications to regions that strongly influence the output decision, such as the traffic light colors or the brake lights of a leading vehicle.
Finally, the semantic guidance leads to a fine preservation of the scene structure in \acro{}'s counterfactuals, achieving both global coherence and high visual quality.

\subsection{Proximity to the Query Image}
\label{sec:expe:expe_sparsity}

We now verify the \emph{proximity} of counterfactuals to query images, as well as the \emph{sparsity} of changes.

\input{tables/sparse_changes}

We first compare \acro{} to previous work with respect to the \textbf{Face Verification Accuracy (FVA)}.
The FVA is the percentage of explanations that preserve the person's identity, as revealed by a cosine similarity above 0.5 between features of the counterfactual and the query.
Following previous works \cite{progressive_exaggeration,dive}, features are computed by a pre-trained re-identification network on \dset{VGGFace2} \cite{vggface2}.
As shown in \autoref{tab:fva}, even if \acro{} is designed for high-quality or complex scenes image classifiers, it reaches high FVA on the low-quality \dset{CelebA} dataset. 
Moreover, \acro{} significantly outperforms PE and DiVE on \dset{CelebAMask-HQ}, showing its ability to scale up to higher image sizes. 
Again, DiVE suffers from the poor capacities of $\beta-$TCVAE to reconstruct high-quality images \autoref{sec:rw}.
To support this claim, we compute the FVA between query images and reconstructions with the $\beta-$TCVAE of DiVE and obtain 45.9\%, which indicates a low reconstruction capacity.

We then measure the sparsity of explanations using the \textbf{Mean Number of Attributes Changed (MNAC)}.
This metric averages the number of facial attributes that differ between the query image and its associated counterfactual explanation.
As \acro{} successfully switches the model's decision almost every time, explanations that obtain a low MNAC are likely to have altered only the necessary elements to build a counterfactual.
Following previous work \cite{dive}, we use an oracle ResNet pretrained on \dset{VGGFace2} \cite{vggface2}, and fine-tuned on 40 attributes provided in \dset{CelebA}/\dset{CelebAMask-HQ}.
As reported in \autoref{tab:mnac}, \acro{} has a lower MNAC than PE and DiVE on both \dset{CelebA} and \dset{CelebAMask-HQ}. 
Conditioning the counterfactual generation on semantic masks helps obtaining small variations that are meaningful enough for the model to switch its decision.
This property makes \acro{} useful in practice and well-suited to explain image classifiers.%

\begin{figure*}[t]
    \centering
    \includegraphics[width=\linewidth]{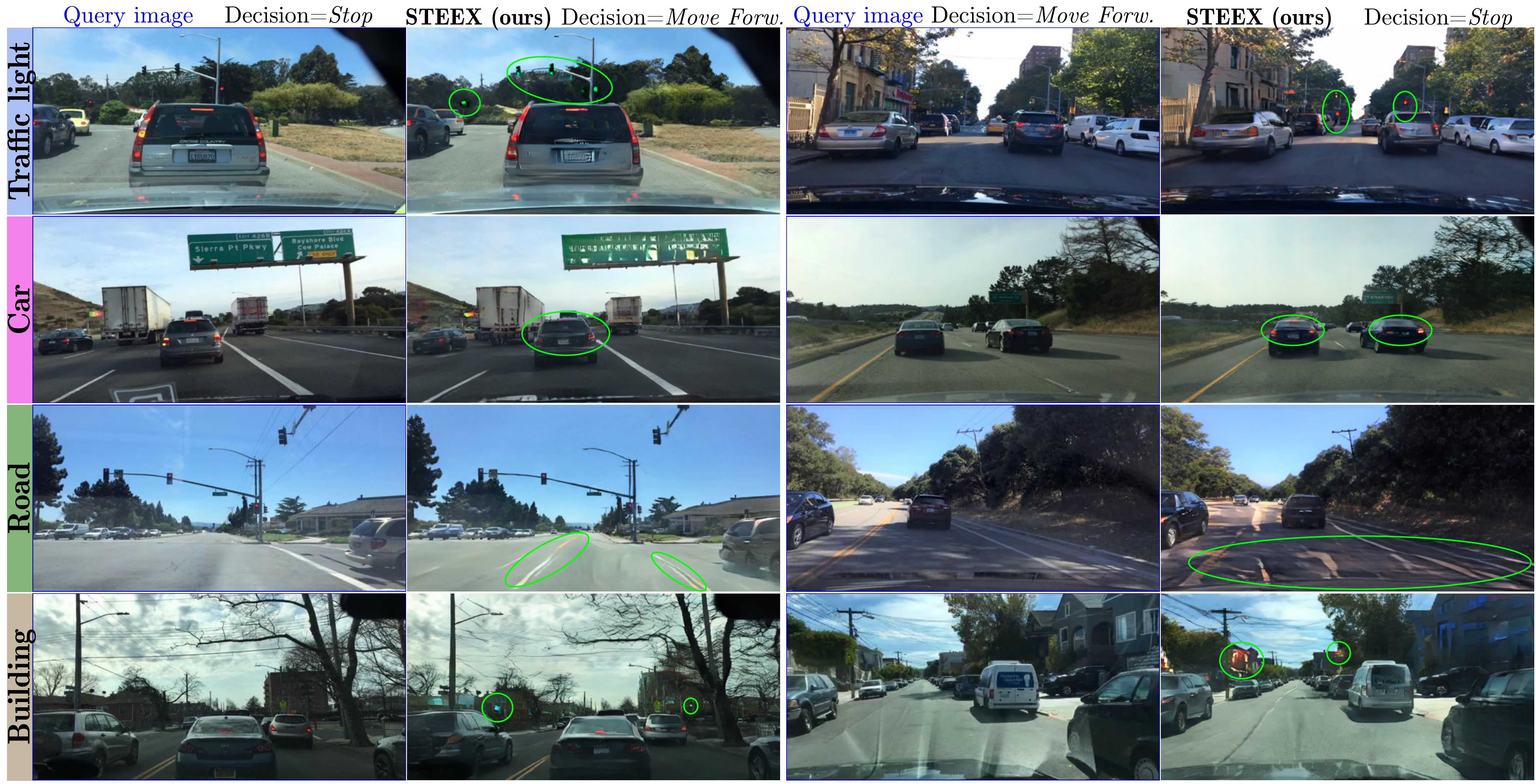}
    \caption{
        \textbf{Semantic region-targeted counterfactual explanations on \dset{BDD100k}.}
        Explanations are generated for a binary classifier trained on the attribute \emph{Move Forward}, at resolution $512 \times 256$. Each row shows explanations where we restrict the optimization process to one specific semantic region, on two examples: one where the model initially goes forward, and one where it initially stops. Significant modifications are highlighted within the green ellipses. 
        Note that even when targeting specific regions, others may still slightly differ from the original image: this is mostly due to small errors in the reconstruction $G(S^I,z^I) \approx x^I$ (more details in the Supplementary)
    }
    \label{fig:bdd_per_class}
\end{figure*}

\subsection{Region-Targeted Counterfactual Explanations}
\label{sec:expe:region_targeted}

As can be seen in Figs. \ref{fig:bigpicture:b} and \ref{fig:bdd_global}, when the query image is complex, the counterfactual explanations can encompass multiple semantic concepts at the same time.
In \autoref{fig:bigpicture:b} for instance, in order to switch the decision of the model to \emph{Move Forward}, the traffic light turns green and the car's brake lights turn off.
It raises ambiguity about how these elements compound to produce the decision.
In other words, ``Are both changes necessary, or changing only one region is sufficient to switch the model's decision?''.

To answer this question, we generate \emph{region-targeted} counterfactual explanations, as explained in \autoref{sec:model:region_targeted}.
In Fig.\,\ref{fig:bigpicture:d}, we observe that targeting the traffic light region can switch the decision of the model, despite the presence of a stopped car blocking the way.
Thereby, region-targeted counterfactuals can help to identify potentially safety-critical issues with the decision model. 

More generally, region-targeted counterfactual explanations empower the user to separately assess how different concepts impact the decision.
We show in \autoref{fig:bdd_per_class} qualitative examples of such region-targeted counterfactual explanations on the \emph{Move Forward} classifier.
On the one hand, we can verify that the decision model relies on cues such as the color of the traffic lights and brake lights of cars, as changing them often successfully switch the decision. 
On the other hand, we discover that changes in the appearance of buildings can flip the model's decision.
Indeed, we see that green or red gleams on facades can fool the decision model into predicting \emph{Move Forward} or \emph{Stop} respectively, suggesting that the model could need further investigation before being safely deployed.

\begin{figure*}[t]
    \centering
    \begin{minipage}[c]{0.6\textwidth}
        \includegraphics[width=\linewidth]{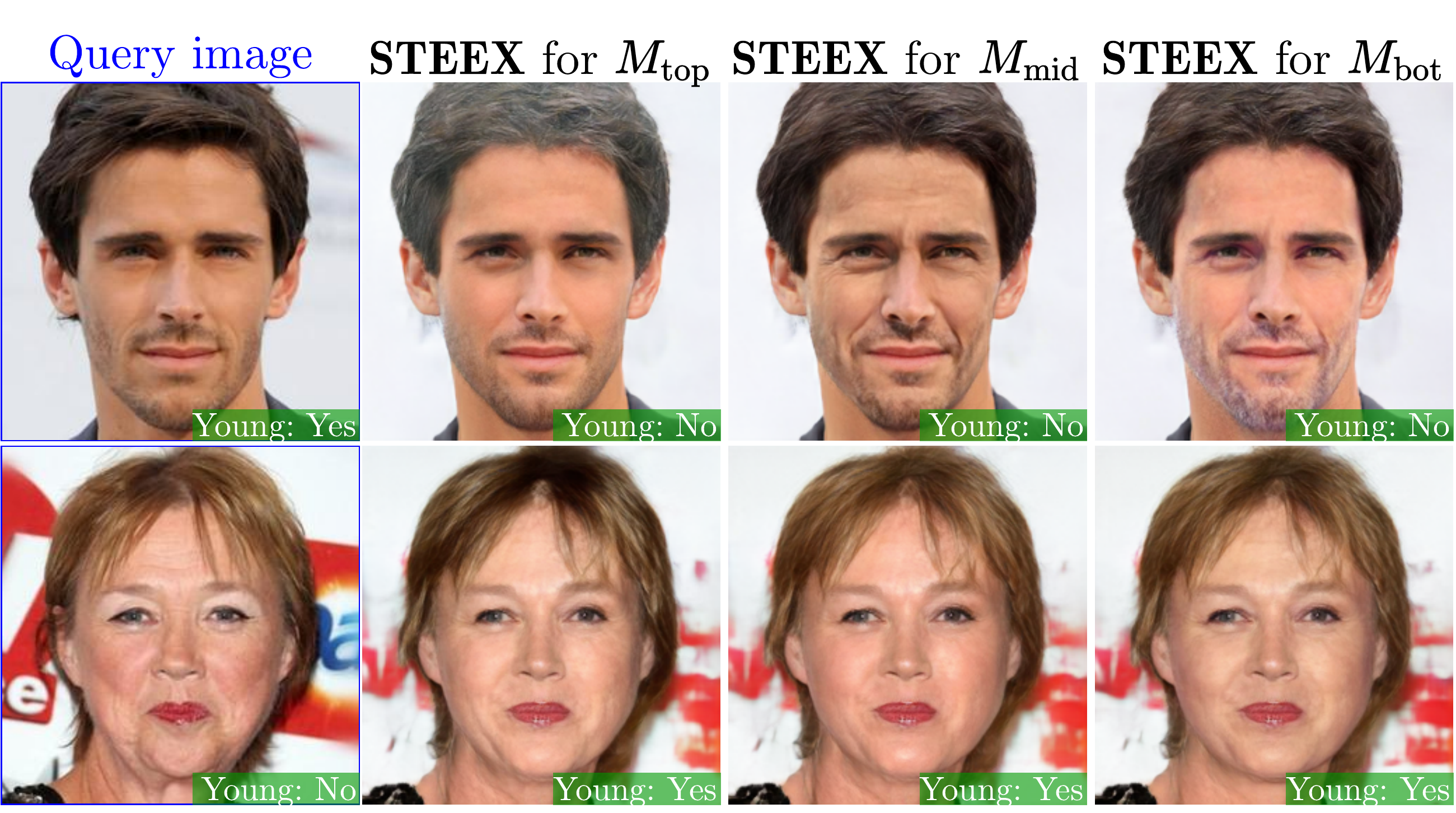}
    \end{minipage}
    \hfill
    \begin{minipage}[c]{0.36\textwidth}
        \caption{
        \textbf{Counterfactual explanations on \dset{CelebAMask-HQ} for three different \emph{Young} classifiers,} namely $M_\text{top}$, $M_\text{mid}$, and $M_\text{bot}$ that respectively only attend to the top, mid, and bottom parts of the image.
        Other examples are available in the Supplementary
        }
        \label{fig:different_decision_models}
    \end{minipage}
\end{figure*}

\subsection{Analyzing Decision Models}
\label{sec:expe:analyzing_decision_models}

An attractive promise of explainable AI is the possibility to detect and characterize biases or malfunctions of explained decision models.
In this section, we investigate how specific are explanations to different decision models and if the explanations can point at the particularity of each model.
In practice, we consider three decision models, namely $M_\text{top}$, $M_\text{mid}$, and $M_\text{bot}$, that were trained on images with masked out pixels except the for the top, middle, and bottom parts of the input respectively.
\autoref{fig:different_decision_models} reports qualitative results, and we can identify that $M_\text{top}$ has based its decisions mainly on the color of the hair, while $M_\text{mid}$ uses the wrinkles on the face, and $M_\text{bot}$ focuses on facial hair and the neck.

\input{tables/decision_models_comparison}

We also measure how much each semantic region has been modified to produce the counterfactual.
Accordingly, we assess the impact of a semantic class $c$ in the decision with the average value of $\| \delta_{z_c} \|^\text{\phantom{2}}_2= \| z_c^I - z_c \|^\text{\phantom{2}}_2$ aggregated over the validation set.
Note that while the absolute values of $\delta_{z_c}$ can be compared across the studied decision models, they cannot be directly compared across different semantic classes, as the $z_c$ can be at different scales for different values of $c$ in the generative model.
To make this comparison in \autoref{tab:decision_model_comparison}, we instead
compute the value of $\delta_{z_c}$ for the target model \emph{relatively} to the average value for all models. The semantic classes of most impact in \autoref{tab:decision_model_comparison} indicate how each decision model is biased towards a specific part of the face and ignores cues that are important for the other models.

\subsection{Ablation Study} 
\label{sec:expe:ablation}
\input{tables/ablation}

We propose an ablation study on \dset{CelebAMask-HQ}, reported in \autoref{tab:ablation}, to assess the role of the distance loss $L_\text{dist}$ and the use of predicted segmentation masks. 

First, we evaluate turning off the distance loss by setting $\lambda=0$, such that the latent codes $z_c$ are no longer constrained to be close to $z_c^I$.
Doing so, for both \emph{Young} and \emph{Smile} classifiers, the FVA and FID of \acro{} degrade significantly, which respectively indicate that the explanation proximity to the real images is deteriorated and that the counterfactuals are less plausible.
The distance loss is thus an essential component for \acro{}.

Second, we investigate if the segmentation network $E_\text{seg}$ is a bottleneck in \acro{}.
To do so, we replace the segmenter's outputs with ground-truth masks and generate counterfactual explanations with these.
The fairly similar scores of both settings indicate that \acro{} works well with inferred layouts.

\section{Conclusion}
In this work, we present \acro{}, a method to generate counterfactual explanations for complex scenes, by steering the generative process using predicted semantics.
To our knowledge, we provide the first framework for complex scenes where numerous elements can affect the decision of the target network.
Experiments on driving scenes and high-quality portraits show the capacity of our method to finely explain deep classification models.
For now, \acro{} is designed to generate explanations that preserve the semantic structure. While we show the merits of this property, future work can consider how, within our framework, to handle operations such as shifting, removing, or adding objects, while keeping the explanation simple to interpret.
Finally, we hope that the setup we propose in \autoref{sec:expe:analyzing_decision_models}, when comparing explanations for multiple decision models with known behaviors, can serve as a basis to measure the interpretability of an explanation method.

\bibliographystyle{splncs04}
\bibliography{macros,biblio}

\newpage 

\appendix

\section{Additional Qualitative Samples}

In this section, we show additional samples of counterfactual explanations generated by \acro{}, for the five classifiers mentioned in the main paper (trained on \dset{CelebA}, \dset{CelebAMask-HQ} and \dset{BDD100k}).

\begin{figure*}
    \centering
    \includegraphics[width=0.8\linewidth]{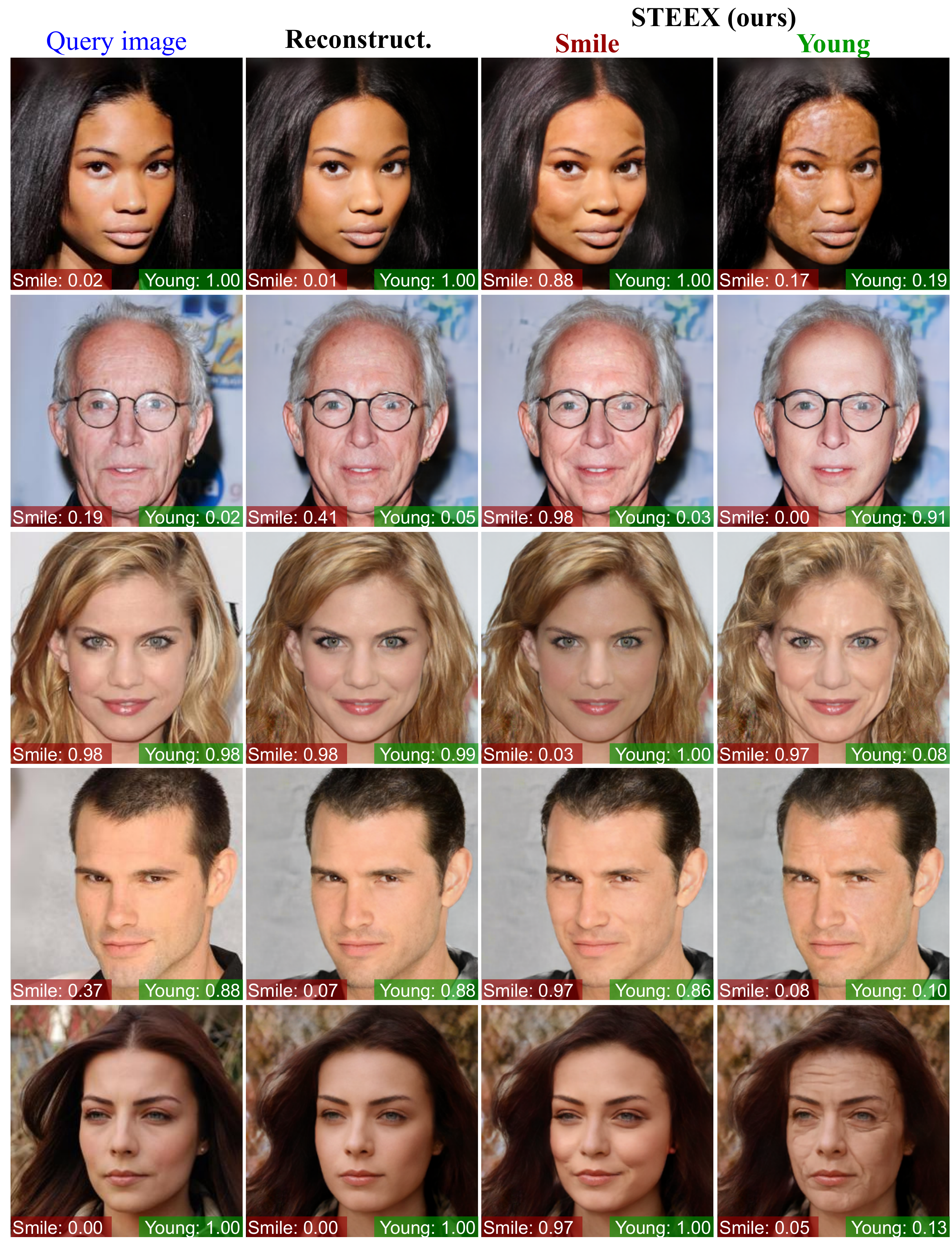}
    \caption{\textbf{Counterfactual explanations and reconstructions on \dset{CelebAMask-HQ} generated by \acro{}}. Explanations are generated for two binary classifiers, on \emph{Smile} and \emph{Young} attributes, at resolution $256 \times 256$.
    Predicted scores are reported at the bottom of each image.
    }
    \label{fig:suppl_celebamhq_a}
\end{figure*}

\begin{figure*}
    \centering
    \includegraphics[width=0.8\linewidth]{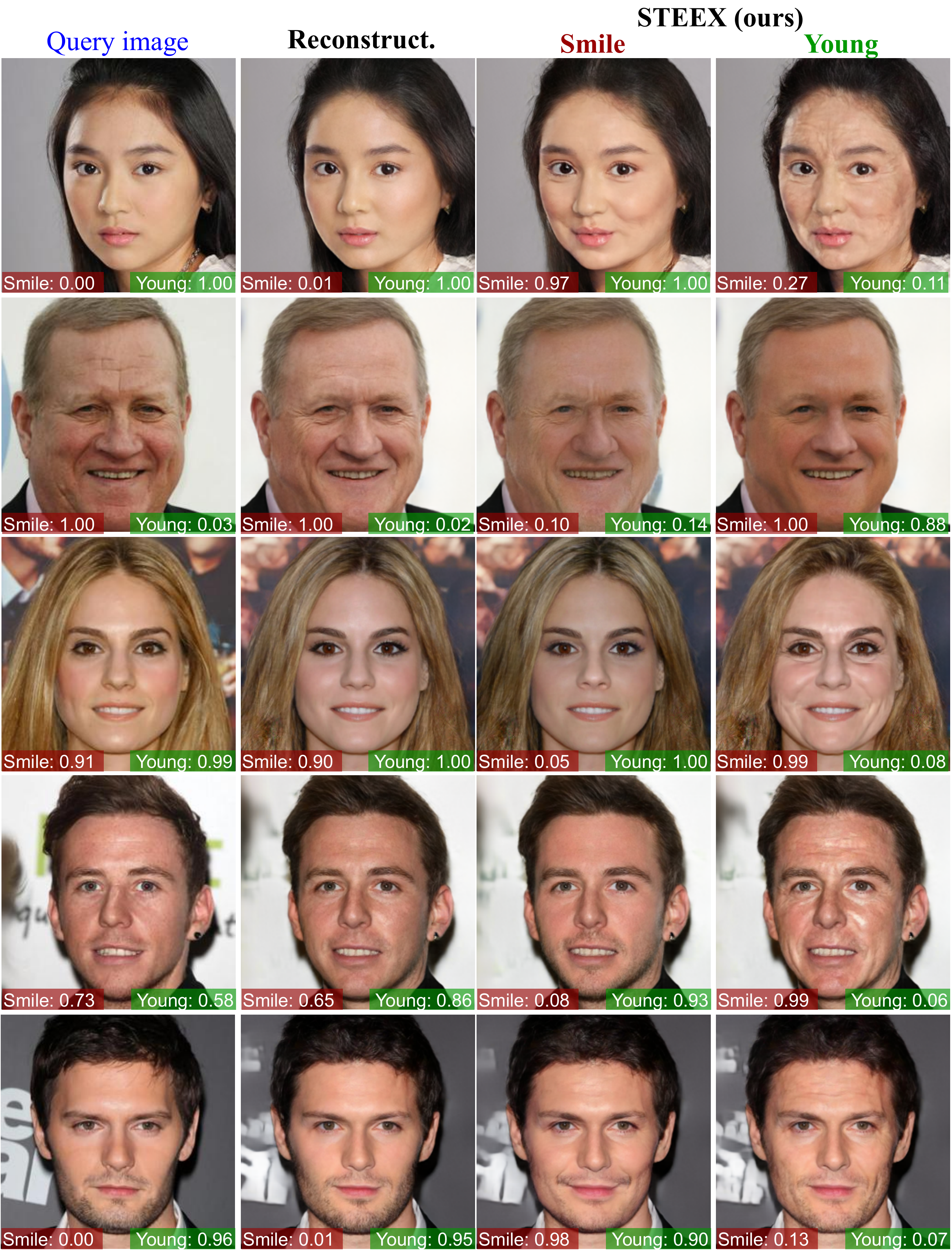}
    \caption{\textbf{Counterfactual explanations and reconstructions on \dset{CelebAMask-HQ} generated by \acro{}}. Explanations are generated for two binary classifiers, on \emph{Smile} and \emph{Young} attributes, at resolution $256 \times 256$.
    Predicted scores are reported at the bottom of each image
    }
    \label{fig:suppl_celebamhq_b}
\end{figure*}

\paragraph{STEEX on \dset{CelebAMask-HQ}.}
In \autoref{fig:suppl_celebamhq_a} and \autoref{fig:suppl_celebamhq_b}, we show samples for the \emph{Smile-} and \emph{Young-} classifiers on the \dset{CelebAMask-HQ} dataset, with images of the size $256 \times 256$.
The modifications found by \acro{} to the images are plausible, understandable and easily traceable by a human due to their sparsity: they are mostly around the mouth for the \emph{Smile}-classifier and on the skin and hair texture for the \emph{Young}-classifier. Note that these explanations are \emph{not} region-targeted, meaning that \acro{} automatically selects the semantics to modify for the explanations.

\begin{figure}[t]
    \centering
    \includegraphics[width=0.6\linewidth]{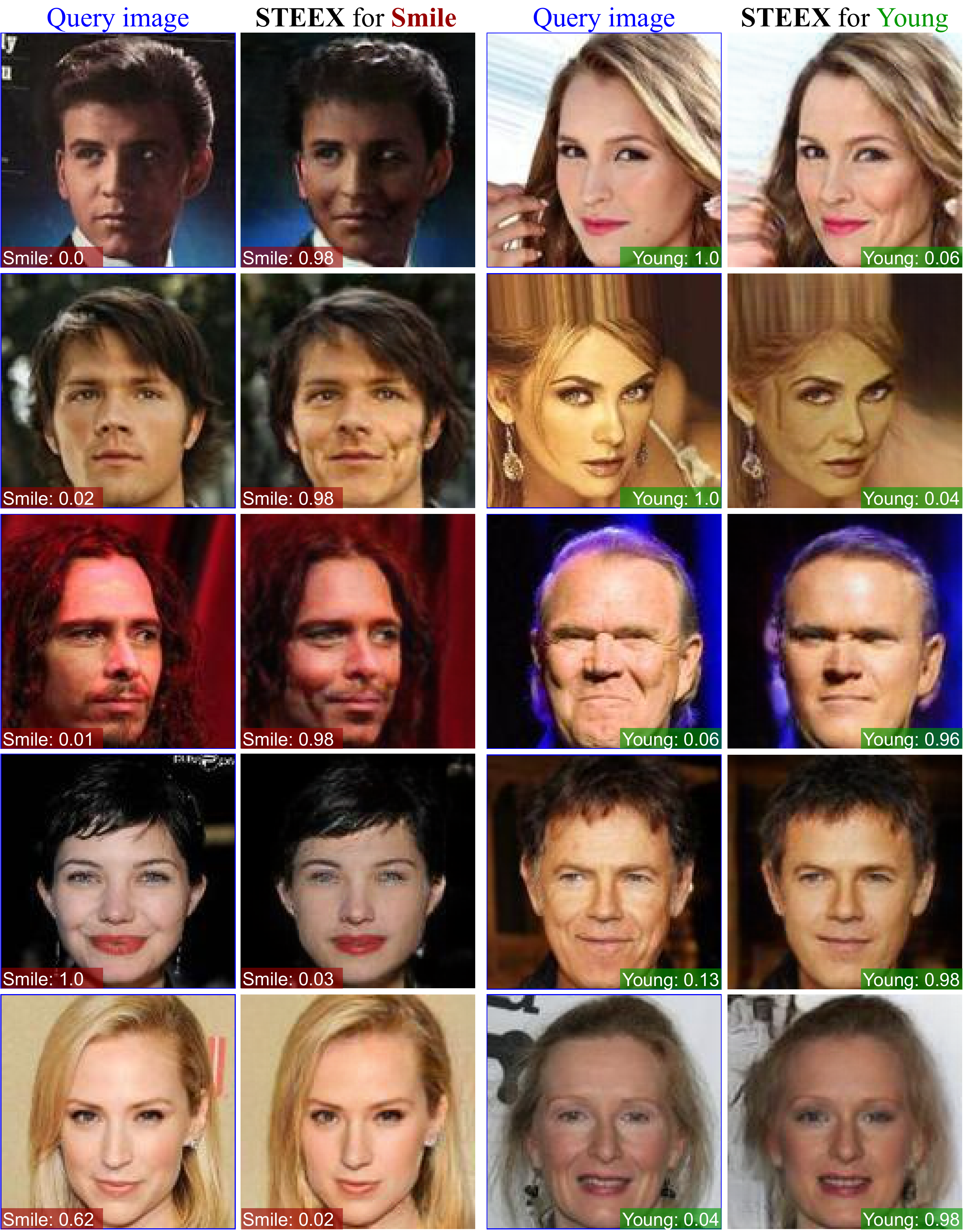}
    \caption{\textbf{Counterfactual explanations on \dset{CelebA} generated by \acro{}}. Explanations are generated for two binary classifiers, on \emph{Smile} and \emph{Young} attributes, at resolution $128 \times 128$.
    Predicted scores are reported at the bottom of each image
    }
    \label{fig:suppl_celeba}
\end{figure}

\paragraph{STEEX on \dset{CelebA}.}
In \autoref{fig:suppl_celeba}, we show samples for the  \emph{Smile-} and \emph{Young-} classifiers on the \dset{CelebA} dataset, with images of the size $128 \times 128$. \acro{} applies both meaningful and sparse modifications to the query images and we can make similar observations as for \dset{CelebAMask-HQ}.

\begin{figure}[t]
    \centering
    \includegraphics[width=0.6\linewidth]{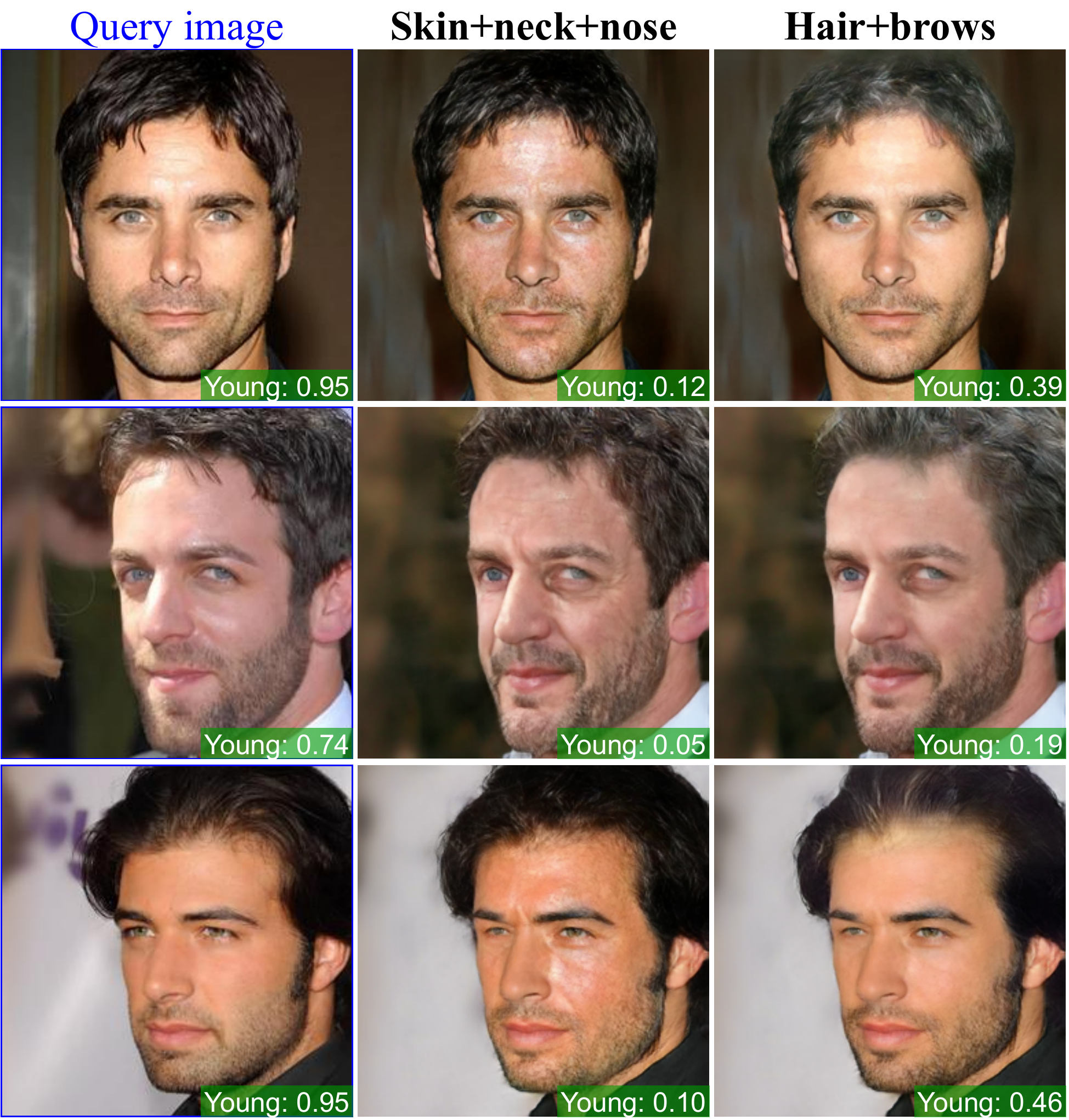}
    \caption{\textbf{Region-targeted counterfactual explanations generated by \acro{} on \dset{CelebAMask-HQ}}. Explanations are generated for a binary classifier on the \emph{Young} attribute. From left to right: query images, counterfactual explanations on the skin, neck and nose, and counterfactual explanations on the hair and eyebrows. On the first set of explanations, \acro{} mostly adds wrinkles, while on the second set, it greys slightly the hair}
    \label{fig:suppl_region_targeted_celebamhq}
\end{figure}

\paragraph{Region-targeted counterfactuals on \dset{CelebAMask-HQ}.}
In \autoref{fig:suppl_region_targeted_celebamhq}, we report examples of region-targeted counterfactual explanations on \dset{CelebAMask-HQ}, for a binary classifier on the attribute \emph{Young}. While the counterfactual explanations targeting the skin regions part mostly add wrinkles to the faces, explanations on the hairy parts (hair and eyebrows) slightly turn them to gray. As skin-targeted counterfactuals are more convincing than hair-targeted counterfactuals, it may indicate that the decision model mostly relies on the skin texture and wrinkles to perform its `\emph{Young}' classification.

\begin{figure*}[t]
    \centering
    \includegraphics[width=\linewidth]{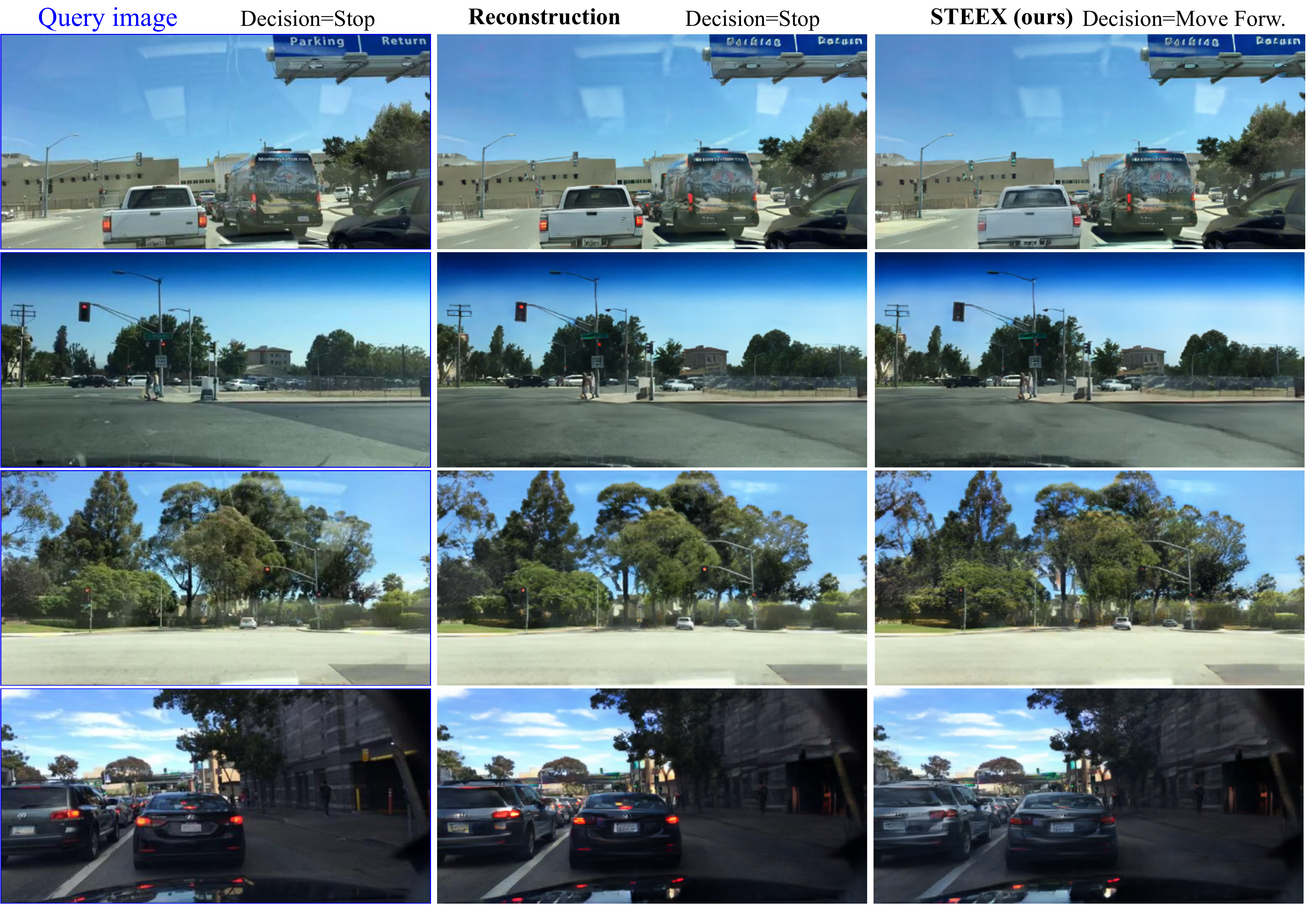}
    \caption{\textbf{Counterfactual explanations on \dset{BDD100k} generated by \acro{}, where the decision model initially predicts `\emph{Stop}'}.
    Explanations are generated for a binary classifier trained with the \dset{BDD-OIA} dataset extension annotated with the attribute \emph{Move forward}. The image resolution is $512 \times 256$
    }
    \label{fig:suppl_bdd_tomoveforw}
\end{figure*}

\begin{figure*}[t]
    \centering
    \includegraphics[width=\linewidth]{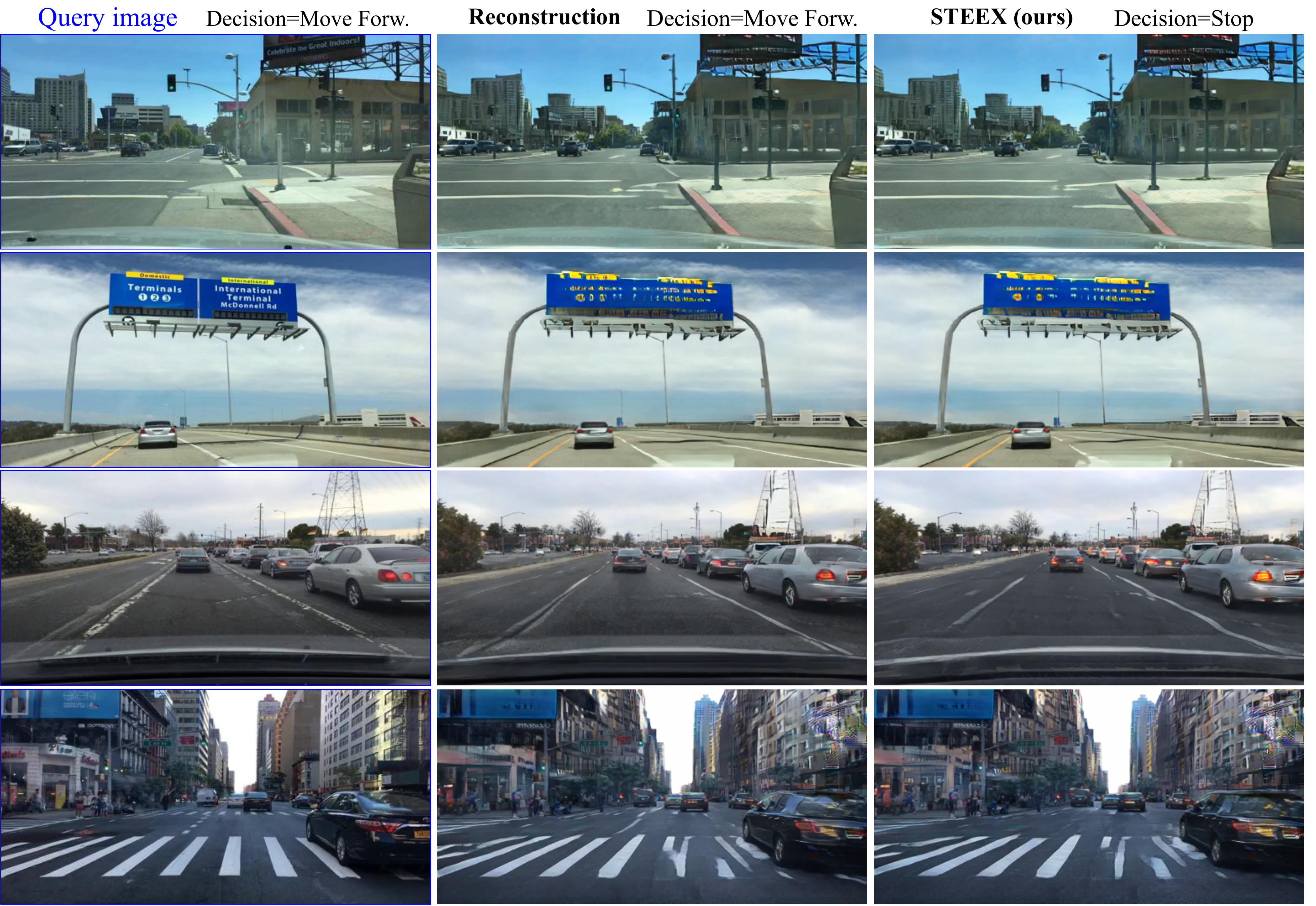}
    \caption{
    \textbf{Counterfactual explanations on \dset{BDD100k} generated by \acro{}, where the decision model initially predicts the `\emph{Move forward}' class}.
    Explanations are generated for a binary classifier trained with the \dset{BDD-OIA} dataset extension annotated with the attribute \emph{Move forward}. The image resolution is $512 \times 256$
    }
    \label{fig:suppl_bdd_tostop}
\end{figure*}

\paragraph{STEEX on \dset{BDD100k}.}
In \autoref{fig:suppl_bdd_tomoveforw} and \autoref{fig:suppl_bdd_tostop}, we show samples for the \textit{Move-forward} classifier on the \dset{BDD100k} dataset, with images of size $512 \times 256$.
To explain `\emph{Stop}' decisions, by providing counterfactual images where the decision model predicts `\emph{Move forward}', several modifications can be observed depending on the image at hand, as reported by \autoref{fig:suppl_bdd_tomoveforw}.
The red light of traffic-lights can fade away (no light at all), or a green light can appear (top image).
Besides, the back brake lights of the front vehicle can fade away as well. Interestingly, we observe on the top image that the brake lights of the front %
vehicle are more impacted than the brake light of the vehicle on the side. This may indicate that the decision model learned to mostly rely on the back lights of the front vehicle and not so much on vehicles of other lanes. 
On the other hand, in \autoref{fig:suppl_bdd_tostop}, to explain `\emph{Move Forward}' decisions, by providing counterfactual images where the decision model predicts `\emph{Stop}', modifications include green traffic lights fading away, and rear brake lights of front cars turning on, as well as slight modification of the road texture which may indicate some spurious correlations learned by the decision model. 

\begin{figure*}[t]
    \centering
    \includegraphics[width=\linewidth]{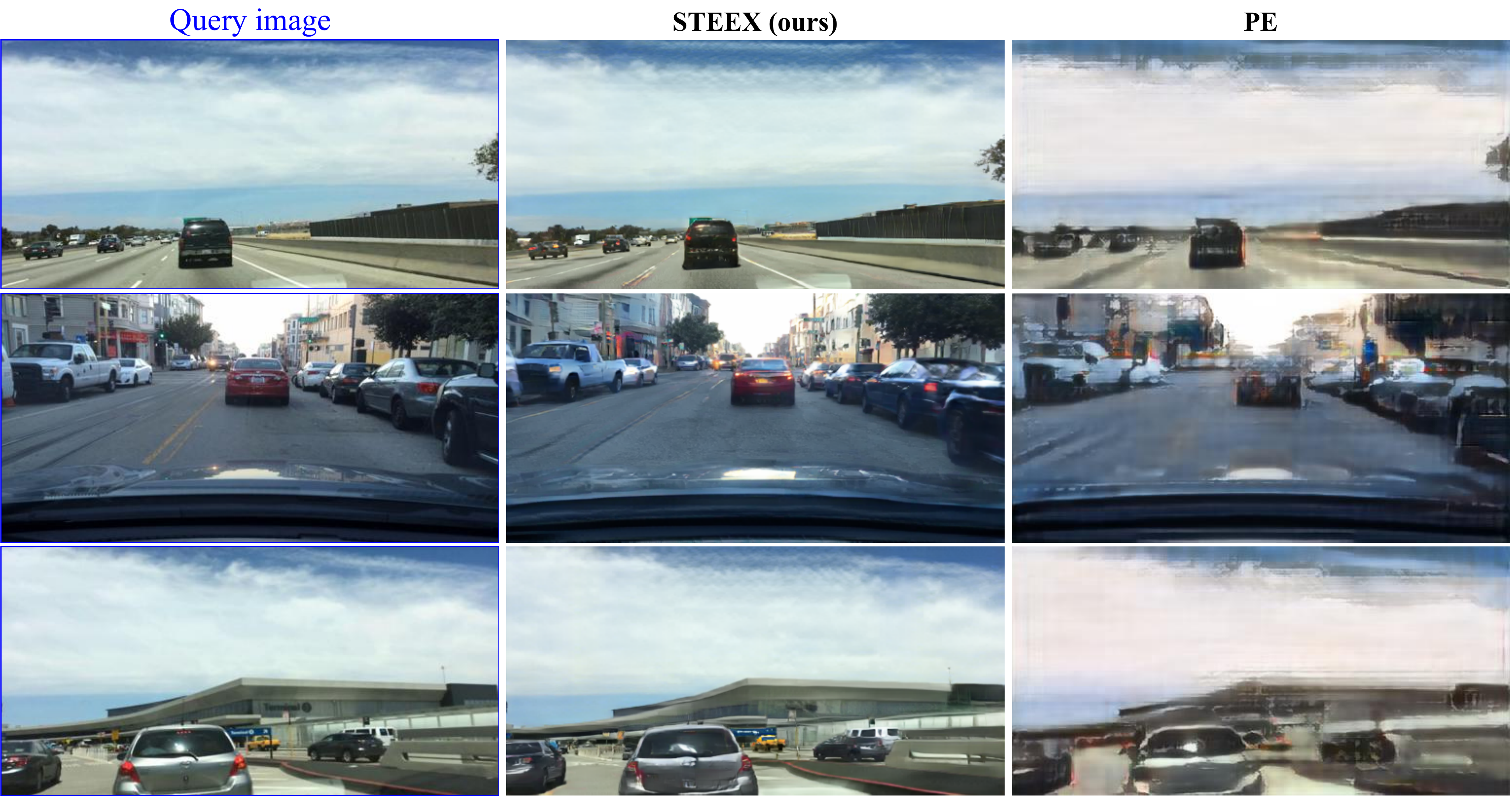}
    \caption{
    \textbf{Counterfactual explanations on \dset{BDD100k} generated by \acro{} compared to explanations generated by Progressive Exaggeration (PE) \cite{progressive_exaggeration}}. All images have a $512 \times 256$ resolution
    }
    \label{fig:suppl_bdd_pe}
\end{figure*}

\paragraph{STEEX vs.\ PE on \dset{BDD100k}.}
In \autoref{fig:suppl_bdd_pe}, we present a comparison between STEEX and PE \cite{progressive_exaggeration} counterfactuals on the same query image for the \emph{Move forward} classifier on \dset{BDD100k} query images.
We observe that counterfactual explanations produced by PE are blurred and, critically, they lose important details of the query image.
On the other hand, \acro{} successfully retrieves the details of the query image while applying plausible meaningful modifications. 
As explained in the main paper, we recall that, despite our best efforts, the adaptation of DiVE \cite{dive} to the driving scene dataset \dset{BDD100k} produces mostly grey images. Indeed, DiVE suffers from the poor capacities of $\beta-$TCVAE to reconstruct high-quality images.

\begin{figure*}[t]
    \centering
    \includegraphics[width=0.8\linewidth]{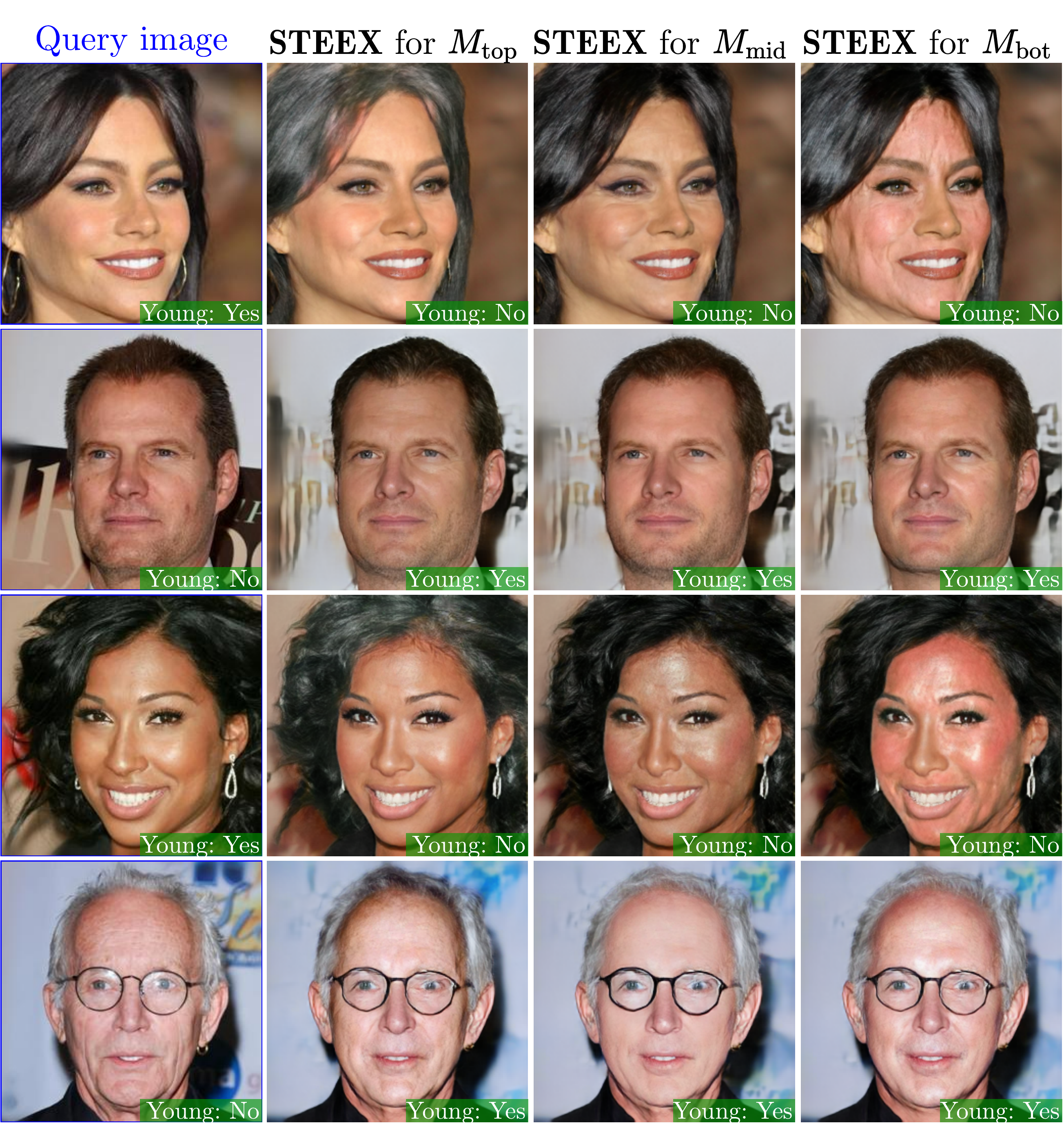}
    \caption{
    \textbf{Counterfactual explanations on \dset{CelebAMask-HQ} for three different \emph{Young} classifiers,} namely $M_\text{top}$, $M_\text{mid}$, and $M_\text{bot}$ that were trained on images where only the top, mid, and bottom parts of images were shown respectively.
    All images have a $256 \times 256$ resolution
    }
    \label{fig:suppl_different_decision_models}
\end{figure*}

\paragraph{STEEX on different decision models on \dset{CelebAMask-HQ}.}
In \autoref{fig:suppl_different_decision_models}, we show additional samples for the three different \emph{Young-} classifiers on the \dset{CelebAMask-HQ} dataset, with images of the size $256 \times 256$.
Modifications found by \acro{} hint at the specificities of each model: we can identify that $M_\text{top}$ has based its decisions mainly on the color of the hair, while $M_\text{mid}$ uses the wrinkles on the face, and $M_\text{bot}$ focuses on facial hair and the neck.

More details about the different models are given in \autoref{sec:suppl_details_analysis_decision_models}.

\section{Reconstruction Quality}
\label{sec:reconstruction}

In this section, we evaluate the impact of the \emph{reconstruction} on the quality and sparsity of the generated counterfactuals.
More precisely, we call `\emph{reconstruction}' the image $G(S^I, z^I)$ generated from the predicted semantic mask $S^I = E_\text{seg}(x^I)$ and the semantic code $z^I = E_z(x^I, S^I)$ obtained on the query image $x^I$. 
Ensuring a good reconstruction quality is crucial.
Indeed, the reconstructed image is the starting point of the optimization towards the counterfactual explanation.
Thus, the reconstructed image must preserve as much as possible the content of the original query image. 
In a way, the quality of the reconstruction gives an upper bound to the quality of the generated counterfactual explanations.

In \autoref{tab:reconstruction_scores}, we present a quantitative evaluation of the quality (FID) and proximity (FVA, MNAC) between the reconstructed images $G(S^I, z^I)$ and the original query images $x^I$, for the three validation datasets. We recall that the reconstruction does not depend on the decision model $M$, but only on the pretrained networks $E_\text{seg}$, $E_z$, and $G$, which are dataset-specific.
In each case, the results are close to the ones reported in Tab. 1 and Tab. 2 of the main paper meaning that the three metrics computed on our counterfactual explanations almost reach the proxy upper bounds.
We can safely argue that our optimization process does not significantly degrade the images, both in terms of perceptual quality and proximity to the image query.
Yet, improving the reconstruction quality, with better pretrained networks $E_\text{seg}$, $E_z$ and $G$ is thus an avenue for a quantitative boost in the results. 

In \autoref{fig:suppl_celebamhq_a}, \autoref{fig:suppl_celebamhq_b}, \autoref{fig:suppl_bdd_tomoveforw} and \autoref{fig:suppl_bdd_tostop}, we show some examples of reconstructions obtained by \acro{} on \dset{CelebAMask-HQ} and \dset{BDD100k}. Overall, a reconstructed image is highly faithful to its query image. However, looking at some close details, we can remark small changes between the query image and its reconstruction from semantics. This slight information loss then propagates on the final counterfactual explanations. Enhancing the reconstruction quality would yield more closeness between the query image and the counterfactual explanation.

\begin{table}[t]
    \centering
    \begin{tabular}{l c c c}
        \toprule
        & \textbf{FID}$\downarrow$ & \textbf{MNAC}$\downarrow$ & \textbf{FVA}$\uparrow$ (\%) \\
        \hline
        \dset{CelebA} & 8.4 & 2.04 & 99.3 \\
        \dset{CelebAMask-HQ} & 21.7 & 3.72 & 99.8 \\
        \dset{BDD100k} & 56.3 & --- & --- \\
        \bottomrule
    \end{tabular}
    \caption{\textbf{Evaluation of the reconstruction quality.} 
    The reconstructed images are obtained with $G(S^I, z^I)$ and their quality is evaluated w.r.t. the original query images $x^I$ with FID, MNAC and FVA metrics, for the three datasets used in this paper
    }
    \label{tab:reconstruction_scores}
\end{table}

\begin{figure*}
    \centering
    \includegraphics[width=\linewidth]{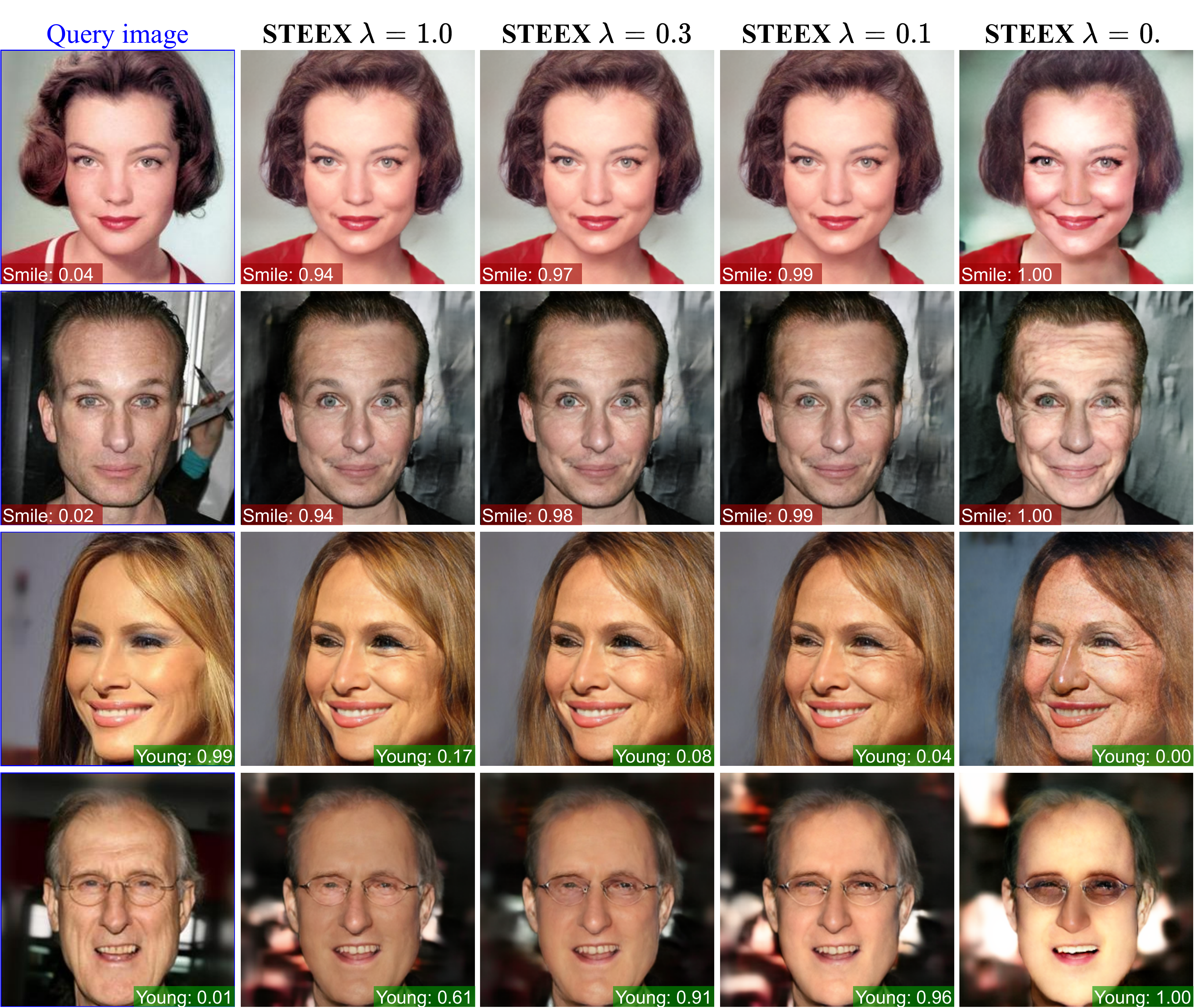}
    \caption{\textbf{Counterfactual explanations with various $\lambda$ generated by \acro{}}. The $\lambda$ parameter balances the contribution of the loss $\mathcal{L}_\text{dist}$ with respect to the one of  $\mathcal{L}_\text{decision}$. When $\lambda$ is high, the decision is `lightly' changed and the counterfactual explanation remains close to the query image. On the contrary, when $\lambda$ is closer to zero, the generated counterfactual explanation is further from the query image and the decision is `heavily' flipped}
    \label{fig:suppl_several_lambda_celebamhq}
\end{figure*}

\section{Details on the analysis of decision models (Sec.\,4.5)}
\label{sec:suppl_details_analysis_decision_models}

The three different classifiers $M_\text{top}$, $M_\text{mid}$, and $M_\text{bot}$, presented in Sec.\,4.5 are trained on modified images of the train set of \dset{CelebAMask-HQ} where all pixels are masked out (with zeros) except for the top, middle, and bottom parts of the image respectively. 
More precisely, $M_\text{top}$ only sees the top 65 pixel rows (out of 256), $M_\text{bot}$ only keeps the bottom 56 pixel rows (out of 256) and $M_\text{mid}$ only sees images where a centered rectangle of size $110\times60$.
The decision model $M_\text{full}$ is the one used for all other experiments, which is trained on unmodified images of the training set of \dset{CelebAMask-HQ}.
Note that the query image from the validation set on which the counterfactual explanation is provided is never modified.
Model accuracies on the Young class are as follow: $M_\text{full}: 89\%$, $M_\text{top}: 83\%$, $M_\text{mid}: 87\%$, $M_\text{bot}: 86\%$.

\section{Technical Details}

\subsection{Pseudo-code}

In Alg.\,\ref{alg:pseudo_code}, we present the pseudo-code to generate a counterfactual explanation for the query image $x^I$ on the model $M$ with our method \acro{}. It assumes that the semantic encoder $E_z$, the semantic segmentation network $E_\text{seg}$ and the generator $G$ have been previously pre-trained. The variable $C$ is used to specify semantic regions in the region-targeted setting. In the general setting, the variable $C$ simply includes all regions of the image.  

\subsection{Selection of the Hyper-parameter $\lambda$}

The hyper-parameter $\lambda$, which balances the respective contributions between the decision loss $L_\text{decision}$ and the distance loss $L_\text{dist}$, was selected as the highest value such that the success-rate was almost perfect ($>99.5 \%$) on the training set of each dataset.
For each of the five decision models, $\lambda=0.3$. With higher values for $\lambda$, the decision is not always flipped. On the other hand, lower values imply that the obtained counterfactual explanation is further from the original query image and the person identity may be lost or more attributes may change. Setting $\lambda = 0$ implies that the distance loss has no contribution in the optimization, meaning that the only objective is the target decision.

We illustrate this in \autoref{fig:suppl_several_lambda_celebamhq}, where we show qualitative results with varying $\lambda$ values. As a lower value for $\lambda$ allows \acro{} to find examples that are more distant to the query image, one can visualize the traits being more and more distorted towards the target decision, in a similar way to the method developed in Progressive Exaggeration (PE) \cite{progressive_exaggeration}. With $\lambda=0$, \ie there is no distance penalty on the generated counterfactuals, images move away from the distribution of natural images, and we cannot consider that they are close enough to the type of images that the decision model $M$ has been trained on, thus loosing the interest of the explanation. Still, it gives insights into the decision mode as it exaggerates important features for the decision model $M$.

\begin{algorithm*}
\caption{\textbf{Pseudo-code for the counterfactual generation by \acro{}.} 
 $x^I$ is the query image and $M$ is the binary decision model.
 $C$ is the subset of regions to be targeted in the region-targeted setting. In the general setting, where counterfactual generation can modify the whole image, $C$ simply includes all semantic regions. 
 $E_\text{seg}$ is a pretrained segmentation network, $E_z$ is a pretrained latent encoder network, $G$ is the generator network. The hyper-parameter $\lambda$ balances the contribution between the two loss terms. $N$ is the number of optimization steps. $l_r$ is the learning rate for the optimization
}\label{alg:pseudo_code} 
\begin{algorithmic}
\Procedure {Generate Counterfactual}{$x^I$, $M$, $C$, $E_\text{seg}$, $E_z$, $G$}
\State $y^I \gets M(x^I)$ \Comment{\textcolor{brown}{\scriptsize Compute the original decision obtained for the query image.}}
\If{$y^I > 0.5$} \Comment{\textcolor{brown}{\scriptsize Get the target counter class $y$ for the counterfactual explanation.}}
    \State $y \gets 0$
\Else
    \State $y \gets 1$
\EndIf
\State $S^I \gets E_\text{seg}(x^I)$ \Comment{\textcolor{brown}{\scriptsize Compute the semantic layout of $x^I$.}}
\State $z^I \gets E_z(x^I, S^I)$ \Comment{\textcolor{brown}{\scriptsize Compute the latent codes for each semantic region.}}
\State $z \gets z^I$ \Comment{\textcolor{brown}{\scriptsize Initialize the latent code of the counterfactual explanation with $z^I$.}}
\For{$i \leftarrow 1$ to $N$} \Comment{\textcolor{brown}{\scriptsize Make $N$ optimization steps.}}
    \State $x \gets G(z, S^I)$ \Comment{\textcolor{brown}{\scriptsize Generate $x$ from the current code $z$, along with $S^I$.}}
    \State $\tilde{y} \gets M(x)$ \Comment{\textcolor{brown}{\scriptsize Compute the model decision on $x$.}}
    \State $L \gets \mathcal{L}(\tilde{y}, y) + \lambda \sum\nolimits_{c \in C} \|z^I_c - z_c\|^2_2$~ \Comment{\textcolor{brown}{\scriptsize Compute global objective.}} %
    \State $z \gets \texttt{ADAM}(z, L, C, l_r)$  \Comment{\textcolor{brown}{\scriptsize Update the code $z$ with one gradient step, only on codes $z_c$\\ \hfill with $c \in C$.}}
\EndFor
\State $x \gets G(z, S^I)$ \Comment{\textcolor{brown}{\scriptsize Compute the final counterfactual explanation.}}
\State \textbf{return} $x$
\EndProcedure
\end{algorithmic}
\end{algorithm*}

\subsection{Licenses}

\paragraph{BDD100k data \cite{bdd}.} \url{https://doc.bdd100k.com/license.html}

\paragraph{BDD100k code.}  BSD 3-Clause License

\paragraph{BDD-OIA data \cite{bdd_oia}.}  No license provided

\paragraph{BDD-OIA code.}  BSD 3-Clause License

\paragraph{CelebA \cite{celeba}.} Agreement to use data on \\ \url{https://mmlab.ie.cuhk.edu.hk/projects/CelebA.html}

\paragraph{CelebAMask-HQ \cite{CelebAMask-HQ}.} Agreement to use data on \\ \url{https://mmlab.ie.cuhk.edu.hk/projects/CelebA/CelebAMask_HQ.html}

\paragraph{SEAN \cite{sean} code.} Creative Commons Attribution-NonCommercial-ShareAlike 4.0 International \url{https://github.com/ZPdesu/SEAN/blob/master/LICENSE.md} %

\paragraph{DiVE \cite{dive} code.} Apache License 2.0 \\ \url{https://github.com/ElementAI/beyond-trivial-explanations/blob/master/LICENSE}

\paragraph{PE \cite{progressive_exaggeration} code.} MIT Licence \url{https://github.com/batmanlab/Explanation_by_Progressive_Exaggeration/blob/master/LICENSE.txt}

\paragraph{DeepLabV3 \cite{deeplabv3} code.} BSD 3-Clause License

\paragraph{Pytorch.} BSD \url{https://github.com/pytorch/pytorch/blob/master/LICENSE}

\end{document}

%% file: bigpicture.tex
\begin{figure}
 \centering
 \begin{subfigure}[h]{0.327\linewidth}
     \centering
     \captionsetup{font=footnotesize}
     \caption{{\fontsize{6}{6}\selectfont Query image}}
     \vspace{-0.2cm}
     \includegraphics[width=\linewidth]{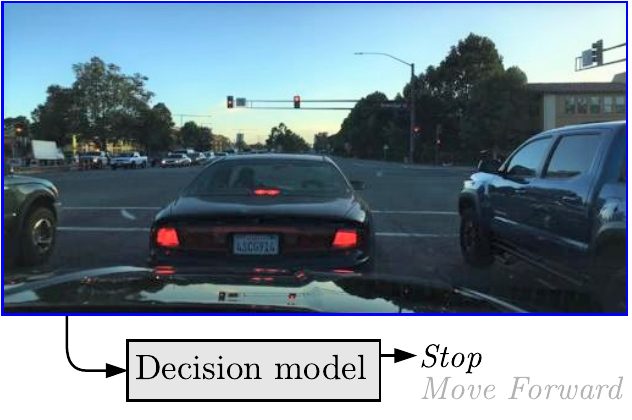}
     \label{fig:bigpicture:a}
 \end{subfigure}
 \hfill
 \begin{subfigure}[h]{0.327\linewidth}
     \centering
     \captionsetup{font=footnotesize}
     \caption{{\fontsize{6}{6}\selectfont Counterfactual explan.\ (CE)}}
     \vspace{-0.2cm}
     \includegraphics[width=\linewidth]{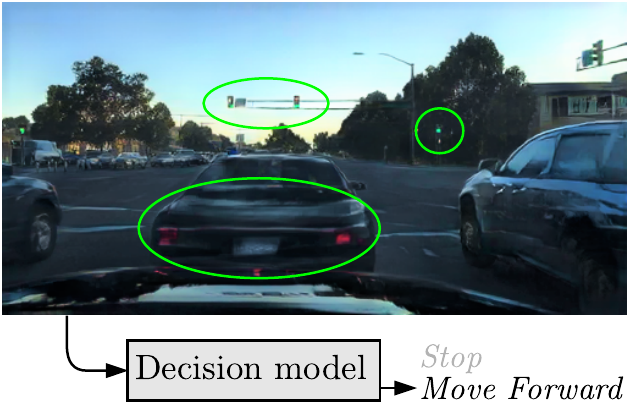}
     \label{fig:bigpicture:b}
 \end{subfigure}
 \hfill
 \begin{subfigure}[h]{0.327\linewidth}
    \centering
    \caption{{\fontsize{6}{6}\selectfont \textbf{Traffic light}-targeted CE}}
     \vspace{-0.2cm}
    \includegraphics[width=\linewidth]{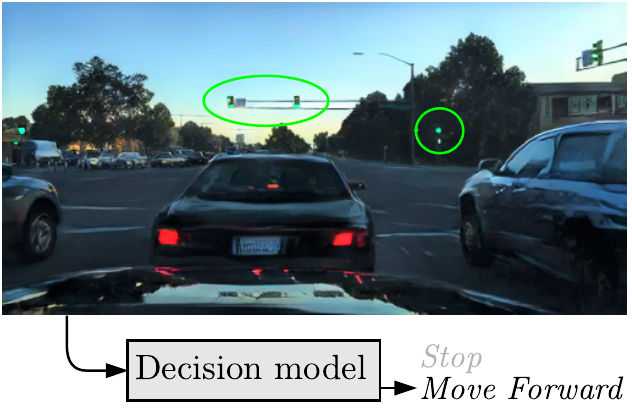}
    \label{fig:bigpicture:d}
 \end{subfigure}
    \vspace{-0.5cm}
    \caption{
    \textbf{Overview of counterfactual explanations generated by our framework \acro{}.}
    Given a trained model and a query image (a), a counterfactual explanation is an answer to the question
    ``\emph{What other image, slightly different and in a meaningful way, would change the model's outcome?}''
    In this example, the `Decision model' is a binary classifier that predicts whether or not it is possible to move forward.
    On top of explaining decisions for large and complex images (b), we propose `region-targeted counterfactual explanations' (c), where produced counterfactual explanations only target specified semantic regions.
    Green ellipses are manually provided to highlight details
	\vspace{-0.8cm}
    }
    \label{fig:bigpicture}
\end{figure}

%% file: overview.tex
\begin{figure*}[t]
    \centering
    \includegraphics[width=\linewidth]{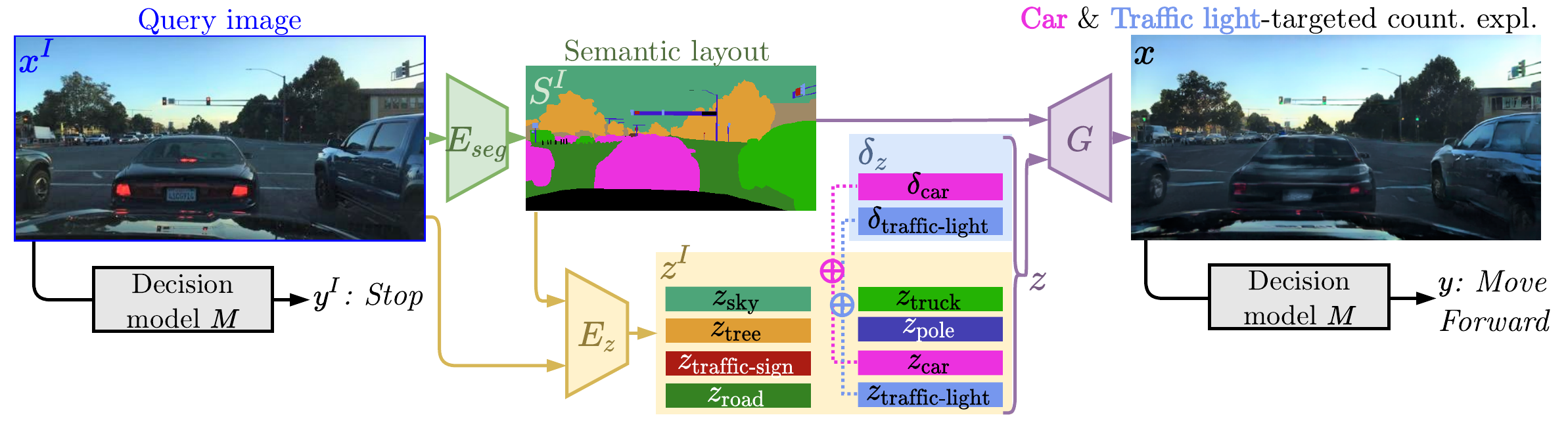}
    \caption{\textbf{Overview of \acro{}.} The query image $x^I$ is first decomposed into a semantic map $S^I$ and $z^I = (z_c)_{c=1}^{N}$, a collection of $N$ semantic embeddings which encode each the aspect of their corresponding semantic category $c$.
    The perturbation $\delta_z$ is optimized such that the generated image $x = G(S^I, z^I + \delta_z)$ is classified as $y$ by the decision model $M$, while staying small.
    As the generator uses the semantic layout $S^I$ of the query image $x^I$, the generated counterfactual explanation $x$ retains the original image structure. 
    The figure specifically illustrates the region-targeted setting, where only the subset \{`car', `traffic light'\} of the semantic style codes is targeted
    }
\label{fig:architecture}
\end{figure*}

%% file: tables/fid_celebahq_bdd.tex
\begin{table}[b]
    \centering
    \caption{\textbf{Perceptual quality, measured with FID$\downarrow$.} Five attribute classifiers are explained, across three datasets.
    Results of PE and DiVE are reported from original papers on \dset{CelebA}.
    For \dset{CelebAMask-HQ} and \dset{BDD100k}, their models are retrained using their code. DiVE does not converge on \dset{BDD100k} %
    }
    \setlength{\tabcolsep}{5pt}
    \begin{tabular}{@{}l rr  rr c@{}}
        \toprule
         \fbox{{FID $\downarrow$}} & \multicolumn{2}{c}{\dset{CelebA}} & \multicolumn{2}{c}{\dset{CelebAM-HQ}} & \dset{BDD100k} \\
         \cmidrule(l){2-3} \cmidrule(l){4-5}
         \cmidrule(l){6-6}& Smile & Young & Smile & Young & Move For. \\
        \midrule
        PE \cite{progressive_exaggeration}& 35.8 & 53.4 & 52.4 & 60.7 & 141.6 \\
        DiVE \cite{dive} & 29.4 & 33.8 & 107.0 & 107.5 & --- \\
        \textbf{STEEX} & \textbf{10.2} & \textbf{11.8} & \textbf{21.9} & \textbf{26.8} & \textbf{~58.8} \\
        \bottomrule
    \end{tabular}
    
    \label{tab:FID_2}
    
\end{table}

%% file: tables/sparse_changes.tex
\begin{table}[b]
    \centering
    \caption{\textbf{Face Verification Accuracy (FVA$\uparrow$) (\%) and Mean Number of Attributes Changed (MNAC$\downarrow$)}, on \dset{CelebA} and \dset{CelebAMask-HQ}. For PE and DiVE, \dset{CelebA} scores come from the original papers, and we re-train their models using official implementations for \dset{CelebAMask-HQ}
    }
    \vspace{0.3cm} 
    \begin{tabular}{@{}l cc c cc@{}}
        \toprule
        \fbox{FVA $\uparrow$} &\multicolumn{2}{c}{\dset{CelebA}} & & \multicolumn{2}{c}{\dset{CelebAM-HQ}} \\
        & Smile & Young & & Smile & Young \\
        \midrule
        PE \cite{progressive_exaggeration} & 85.3 & 72.2 & & 79.8 & 76.2 \\
        DiVE \cite{dive} &\textbf{97.3} & \textbf{98.2} & & 35.7 & 32.3\\
        \textbf{\acro{}} & 96.9 & 97.5 & & \textbf{97.6} & \textbf{96.0} \\
        \bottomrule
        \label{tab:fva}
    \end{tabular}
    \hfill
    \begin{tabular}{@{}l cc c cc@{}}
        \toprule
        \fbox{MNAC $\downarrow$} &\multicolumn{2}{c}{\dset{CelebA}} & & \multicolumn{2}{c}{\dset{CelebAM-HQ}} \\
        & Smile & Young & & Smile & Young \\
        \midrule
        PE \cite{progressive_exaggeration} & --- & 3.74 & & 7.71 & 8.51 \\
        DiVE \cite{dive} & --- & 4.58 & & 7.41 & 6.76 \\
        \textbf{\acro{}} & \textbf{4.11} & \textbf{3.44} & & \textbf{5.27} & \textbf{5.63} \\
        \bottomrule
        \label{tab:mnac}
    \end{tabular}
\end{table}

%% file: tables/decision_models_comparison.tex
\begin{table}[b]
    \centering
    \caption{\textbf{Most and least impactful semantic classes for a decision model relatively to others.} The impact of a class for a given model has been determined as the average value of $\| \delta_{z_c} \|^\text{\phantom{2}}_2= \| z_c^I - z_c \|^\text{\phantom{2}}_2$ for each semantic class $c$, relatively to the same value averaged for other models
    } 
    \begin{tabular}{@{}l c c l c l@{}}
        \toprule
        Model & & & Most impactful & & Least impactful \\ 
        \hline
        $M_\text{top}$ & & & hat, hair, background & & necklace, eyes, lips \\
        $M_\text{mid}$ & & & nose, glasses, eyes & & necklace, neck, hat \\ 
        $M_\text{bot}$ & & & neck, necklace, cloth & & eyes, brows, glasses \\
        \bottomrule
    \end{tabular}
    \label{tab:decision_model_comparison}
\end{table}

%% file: tables/ablation.tex
\begin{table}[t]
    \centering
    \caption{\textbf{Ablation study} measuring the role of the distance loss $L_\text{dist}$ in \autoref{eq:semantic-conterfactual} and upper bound results that would be achieved with ground-truth segmentation masks\\
    }

    \begin{tabular}{@{}l cc c cc@{}}
        \toprule
        & \multicolumn{2}{c}{Smile} & & \multicolumn{2}{c}{Young} \\
        & FID $\downarrow$  & FVA $\uparrow$ & & FID $\downarrow$ & FVA $\uparrow$\\
        \midrule
        \textbf{\acro{}} & 21.9 & 97.6 & & 26.8 & 96.0 \\
        \hspace{0.4cm} without $L_\text{dist}$ & 29.7 & 65.2 & & 45.7 & 37.0 \\
        \hspace{0.4cm} with ground-truth segmentation & 21.2 & 98.9 & & 25.7 & 98.2 \\
        \bottomrule
    \end{tabular}
    
    \label{tab:ablation}
    
\end{table}

%% file: main_arxiv.bbl
\begin{thebibliography}{10}
\providecommand{\url}[1]{\texttt{#1}}
\providecommand{\urlprefix}{URL }
\providecommand{\doi}[1]{https://doi.org/#1}

\bibitem{xai_peeking_inside_black_box}
Adadi, A., Berrada, M.: Peeking inside the black-box: {A} survey on explainable
  artificial intelligence {(XAI)}. {IEEE} Access  (2018)

\bibitem{lrp}
Bach, S., Binder, A., Montavon, G., Klauschen, F., M{\"u}ller, K.R., Samek, W.:
  On pixel-wise explanations for non-linear classifier decisions by layer-wise
  relevance propagation. PloS one  (2015)

\bibitem{beaudouin2020identifying}
Beaudouin, V., Bloch, I., Bounie, D., Cl{\'{e}}men{\c{c}}on, S.,
  d'Alch{\'{e}}{-}Buc, F., Eagan, J., Maxwell, W., Mozharovskyi, P., Parekh,
  J.: Flexible and context-specific {AI} explainability: {A} multidisciplinary
  approach. CoRR  \textbf{abs/2003.07703} (2020)

\bibitem{visualbackprop}
Bojarski, M., Choromanska, A., Choromanski, K., Firner, B., Ackel, L.J.,
  Muller, U., Yeres, P., Zieba, K.: Visualbackprop: Efficient visualization of
  cnns for autonomous driving. In: ICRA (2018)

\bibitem{csgan}
Bora, A., Jalal, A., Price, E., Dimakis, A.G.: Compressed sensing using
  generative models. In: ICML (2017)

\bibitem{why_ce_produce_ae}
Browne, K., Swift, B.: Semantics and explanation: why counterfactual
  explanations produce adversarial examples in deep neural networks. CoRR
  \textbf{abs/2012.10076} (2020)

\bibitem{vggface2}
Cao, Q., Shen, L., Xie, W., Parkhi, O.M., Zisserman, A.: Vggface2: {A} dataset
  for recognising faces across pose and age. In: FG (2018)

\bibitem{explaining_classifiers_counterfactual}
Chang, C., Creager, E., Goldenberg, A., Duvenaud, D.: Explaining image
  classifiers by counterfactual generation. In: ICLR (2019)

\bibitem{this_looks_like_that}
Chen, C., Li, O., Tao, D., Barnett, A., Rudin, C., Su, J.: This looks like
  that: Deep learning for interpretable image recognition. In: NeurIPS (2019)

\bibitem{deeplabv3}
Chen, L., Papandreou, G., Schroff, F., Adam, H.: Rethinking atrous convolution
  for semantic image segmentation. CoRR  \textbf{abs/1706.05587} (2017)

\bibitem{beta_tcvae}
Chen, R.T.Q., Li, X., Grosse, R., Duvenaud, D.: Isolating sources of
  disentanglement in variational autoencoders. In: NeurIPS (2018)

\bibitem{opportunities_and_challenges_in_xai_survey}
Das, A., Rad, P.: Opportunities and challenges in explainable artificial
  intelligence {(XAI):} {A} survey. CoRR  (2020)

\bibitem{meaningful_perturbation}
Fong, R.C., Vedaldi, A.: Interpretable explanations of black boxes by
  meaningful perturbation. In: ICCV (2017)

\bibitem{ce_ae_common_grounds}
Freiesleben, T.: Counterfactual explanations {\&} adversarial examples - common
  grounds, essential differences, and potential transfers. CoRR
  \textbf{abs/2009.05487} (2020)

\bibitem{soft_decision_tree}
Frosst, N., Hinton, G.E.: Distilling a neural network into a soft decision
  tree. In: Workshop on Comprehensibility and Explanation in {AI} and {ML}
  @AI*IA (2017)

\bibitem{GilpinBYBSK18}
Gilpin, L.H., Bau, D., Yuan, B.Z., Bajwa, A., Specter, M., Kagal, L.:
  Explaining explanations: An overview of interpretability of machine learning.
  In: DSSA (2018)

\bibitem{gan2016}
Goodfellow, I.J., Pouget{-}Abadie, J., Mirza, M., Xu, B., Warde{-}Farley, D.,
  Ozair, S., Courville, A.C., Bengio, Y.: Generative adversarial nets. In:
  Ghahramani, Z., Welling, M., Cortes, C., Lawrence, N.D., Weinberger, K.Q.
  (eds.) NeurIPS (2014)

\bibitem{adversarial2015}
Goodfellow, I.J., Shlens, J., Szegedy, C.: Explaining and harnessing
  adversarial examples. In: ICLR (2015)

\bibitem{counterfactual_visual_explanations}
Goyal, Y., Wu, Z., Ernst, J., Batra, D., Parikh, D., Lee, S.: Counterfactual
  visual explanations. In: ICML (2019)

\bibitem{causal_learning_autoencoded_activations}
Harradon, M., Druce, J., Ruttenberg, B.E.: Causal learning and explanation of
  deep neural networks via autoencoded activations. CoRR  (2018)

\bibitem{grounding_visual_explanations}
Hendricks, L.A., Hu, R., Darrell, T., Akata, Z.: Grounding visual explanations.
  In: Ferrari, V., Hebert, M., Sminchisescu, C., Weiss, Y. (eds.) ECCV (2018)

\bibitem{fid}
Heusel, M., Ramsauer, H., Unterthiner, T., Nessler, B., Hochreiter, S.: Gans
  trained by a two time-scale update rule converge to a local nash equilibrium.
  In: NeurIPS (2017)

\bibitem{densenet}
Huang, G., Liu, Z., van~der Maaten, L., Weinberger, K.Q.: Densely connected
  convolutional networks. In: CVPR (2017)

\bibitem{adam}
Kingma, D.P., Ba, J.: Adam: {A} method for stochastic optimization. In: ICLR
  (2015)

\bibitem{stylex}
Lang, O., Gandelsman, Y., Yarom, M., Wald, Y., Elidan, G., Hassidim, A.,
  Freeman, W.T., Isola, P., Globerson, A., Irani, M., Mosseri, I.: Explaining
  in style: Training a {GAN} to explain a classifier in stylespace. In: ICCV
  (2021)

\bibitem{CelebAMask-HQ}
Lee, C.H., Liu, Z., Wu, L., Luo, P.: Maskgan: Towards diverse and interactive
  facial image manipulation. In: CVPR (2020)

\bibitem{high_fidelity_disentangled}
Lee, W., Kim, D., Hong, S., Lee, H.: High-fidelity synthesis with disentangled
  representation. In: ECCV (2020)

\bibitem{discoverbiasedattr2021}
Li, Z., Xu, C.: Discover the {{Unknown Biased Attribute}} of an {{Image
  Classifier}}. In: The {{IEEE International Conference}} on {{Computer
  Vision}} ({{ICCV}}) (2021)

\bibitem{celeba}
Liu, Z., Luo, P., Wang, X., Tang, X.: Deep learning face attributes in the
  wild. In: ICCV (2015)

\bibitem{shap}
Lundberg, S.M., Lee, S.: A unified approach to interpreting model predictions.
  In: NeurIPS (2017)

\bibitem{deepfool}
Moosavi{-}Dezfooli, S., Fawzi, A., Frossard, P.: Deepfool: {A} simple and
  accurate method to fool deep neural networks. In: CVPR (2016)

\bibitem{spade}
Park, T., Liu, M., Wang, T., Zhu, J.: Semantic image synthesis with
  spatially-adaptive normalization. In: CVPR (2019)

\bibitem{connections_ce_ae}
Pawelczyk, M., Joshi, S., Agarwal, C., Upadhyay, S., Lakkaraju, H.: On the
  connections between counterfactual explanations and adversarial examples.
  CoRR  \textbf{abs/2106.09992} (2021)

\bibitem{there_and_back_again}
Rebuffi, S., Fong, R., Ji, X., Vedaldi, A.: There and back again: Revisiting
  backpropagation saliency methods. In: CVPR (2020)

\bibitem{lime}
Ribeiro, M.T., Singh, S., Guestrin, C.: "why should {I} trust you?": Explaining
  the predictions of any classifier. In: SIGKDD (2016)

\bibitem{dive}
Rodr{\'{\i}}guez, P., Caccia, M., Lacoste, A., Zamparo, L., Laradji, I.H.,
  Charlin, L., V{\'{a}}zquez, D.: Beyond trivial counterfactual explanations
  with diverse valuable explanations. In: ICCV (2021)

\bibitem{oasis}
Sch{\"{o}}nfeld, E., Sushko, V., Zhang, D., Gall, J., Schiele, B., Khoreva, A.:
  You only need adversarial supervision for semantic image synthesis. In: ICLR
  (2021)

\bibitem{gradcam}
Selvaraju, R.R., Cogswell, M., Das, A., Vedantam, R., Parikh, D., Batra, D.:
  Grad-cam: Visual explanations from deep networks via gradient-based
  localization. In: ICCV (2017)

\bibitem{shen2020explain}
Shen, Y., Jiang, S., Chen, Y., Yang, E., Jin, X., Fan, Y., Campbell, K.D.: To
  explain or not to explain: {A} study on the necessity of explanations for
  autonomous vehicles. CoRR  (2020)

\bibitem{deeplift}
Shrikumar, A., Greenside, P., Kundaje, A.: Learning important features through
  propagating activation differences. In: ICML (2017)

\bibitem{progressive_exaggeration}
Singla, S., Pollack, B., Chen, J., Batmanghelich, K.: Explanation by
  progressive exaggeration. In: ICLR (2020)

\bibitem{improving_reconstruction_disentangled}
Srivastava, A., Bansal, Y., Ding, Y., Hurwitz, C., Xu, K., Egger, B.,
  Sattigeri, P., Tenenbaum, J., Cox, D.D., Gutfreund, D.: Improving the
  reconstruction of disentangled representation learners via multi-stage
  modelling. CoRR  \textbf{abs/2010.13187} (2020)

\bibitem{integrated_gradients}
Sundararajan, M., Taly, A., Yan, Q.: Axiomatic attribution for deep networks.
  In: ICML (2017)

\bibitem{adversarial2014}
Szegedy, C., Zaremba, W., Sutskever, I., Bruna, J., Erhan, D., Goodfellow,
  I.J., Fergus, R.: Intriguing properties of neural networks. In: ICLR (2014)

\bibitem{deeptest}
Tian, Y., Pei, K., Jana, S., Ray, B.: Deeptest: automated testing of
  deep-neural-network-driven autonomous cars. In: ICSE (2018)

\bibitem{deep_image_prior}
Ulyanov, D., Vedaldi, A., Lempitsky, V.S.: Deep image prior. IJCV  (2020)

\bibitem{counterfactual_explanations_review}
Verma, S., Dickerson, J.P., Hines, K.: Counterfactual explanations for machine
  learning: {A} review. CoRR  \textbf{abs/2010.10596} (2020)

\bibitem{wachter2017counterfactual}
Wachter, S., Mittelstadt, B., Russell, C.: Counterfactual explanations without
  opening the black box: Automated decisions and the gdpr. Harvard Journal of
  Law \& Technology  (2017)

\bibitem{interpretable_fine_grained_expl}
Wagner, J., K{\"{o}}hler, J.M., Gindele, T., Hetzel, L., Wiedemer, J.T.,
  Behnke, S.: Interpretable and fine-grained visual explanations for
  convolutional neural networks. In: CVPR (2019)

\bibitem{scout}
Wang, P., Vasconcelos, N.: {SCOUT:} self-aware discriminant counterfactual
  explanations. In: CVPR (2020)

\bibitem{bdd_oia}
Xu, Y., Yang, X., Gong, L., Lin, H.C., Wu, T.Y., Li, Y., Vasconcelos, N.:
  Explainable object-induced action decision for autonomous vehicles. In: CVPR
  (2020)

\bibitem{bdd}
Yu, F., Chen, H., Wang, X., Xian, W., Chen, Y., Liu, F., Madhavan, V., Darrell,
  T.: {BDD100K:} {A} diverse driving dataset for heterogeneous multitask
  learning. In: CVPR (2020)

\bibitem{xai_driving_survey}
Zablocki, {\'{E}}., Ben{-}Younes, H., P{\'{e}}rez, P., Cord, M.: Explainability
  of vision-based autonomous driving systems: Review and challenges. CoRR
  \textbf{abs/2101.05307} (2021)

\bibitem{deconvnet}
Zeiler, M.D., Fergus, R.: Visualizing and understanding convolutional networks.
  In: ECCV (2014)

\bibitem{deeproad}
Zhang, M., Zhang, Y., Zhang, L., Liu, C., Khurshid, S.: Deeproad: Gan-based
  metamorphic testing and input validation framework for autonomous driving
  systems. In: IEEE ASE (2018)

\bibitem{trusthci20}
Zhang, Q., Yang, X.J., Robert, L.P.: Expectations and trust in automated
  vehicles. In: CHI (2020)

\bibitem{interpretable_cnn}
Zhang, Q., Wu, Y.N., Zhu, S.: Interpretable convolutional neural networks. In:
  CVPR (2018)

\bibitem{object_detectors_emerge_in_deep_scene}
Zhou, B., Khosla, A., Lapedriza, {\`{A}}., Oliva, A., Torralba, A.: Object
  detectors emerge in deep scene cnns. In: ICLR (2015)

\bibitem{cam}
Zhou, B., Khosla, A., Lapedriza, {\`{A}}., Oliva, A., Torralba, A.: Learning
  deep features for discriminative localization. In: CVPR (2016)

\bibitem{sean}
Zhu, P., Abdal, R., Qin, Y., Wonka, P.: {SEAN:} image synthesis with semantic
  region-adaptive normalization. In: CVPR (2020)

\end{thebibliography}
